%% file: writeup.tex
\documentclass{article}

\usepackage[final]{nips_2017}

\usepackage{graphicx}
\usepackage[utf8]{inputenc} 
\usepackage[T1]{fontenc}    
\usepackage{hyperref}       
\usepackage{url}            
\usepackage{booktabs}       
\usepackage{amsfonts}       
\usepackage{nicefrac}       
\usepackage{microtype}      
\usepackage{array}          
\usepackage{placeins}
\usepackage{caption}
\usepackage{tikz}
\usepackage{pgfplots}
\newcolumntype{P}[1]{>{\centering\arraybackslash}p{#1}} 

\title{Semantic Classification of Tabular Datasets via Character-Level Convolutional Neural Networks}

\author{
  Paul Azunre \\
  \And
  Craig Corcoran \\
  \And
  Numa Dhamani\\
  \AND
  Jeffrey Gleason\\
  \And
  Garrett Honke \\
  \And
  David Sullivan \\
  \AND
  Rebecca Ruppel\\
  \And
  Sandeep Verma \\
  New Knowledge\\
  Austin, TX\\
  \texttt{\{first.name\}@newknowledge.io}\\
  \And
  Jonathon Morgan\\
}

\begin{document}

\maketitle

\begin{abstract}
  A character-level convolutional neural network (CNN) motivated by applications in ``automated machine learning'' (AutoML) is proposed to semantically classify columns in tabular data. Simulated data containing a set of base classes is first used to learn an initial set of weights. Hand-labeled data from the CKAN repository is then used in a transfer-learning paradigm to adapt the initial weights to a more sophisticated representation of the problem (e.g., including more classes). In doing so, realistic data imperfections are learned and the set of classes handled can be expanded from the base set with reduced labeled data and computing power requirements. Results show the effectiveness and flexibility of this approach in three diverse domains: semantic classification of tabular data, age prediction from social media posts, and email spam classification. In addition to providing further evidence of the effectiveness of transfer learning in natural language processing (NLP), our experiments suggest that analyzing the semantic structure of language at the character level without additional metadata---i.e., network structure, headers, etc.---can produce competitive accuracy for type classification, spam classification, and social media age prediction. We present our open-source toolkit SIMON, an acronym for Semantic Inference for the Modeling of ONtologies, which implements this approach in a user-friendly and scalable/parallelizable fashion.  
\end{abstract}

\section{Introduction}

Text classification---the process of assigning a set of tags or predefined categories to free-text---is a fundamental task for natural language processing with widespread applications ranging from topic labeling and sentiment analysis to spam detection. Traditional text classification methods based on simple statistics of word-ordered combinations have been known to perform the best when applied to a closely-defined domain, but are limited in their knowledge of words and the semantic structure of language \citep{1, 2, 3}. Recent research shows that deep learning systems applied to various large-scale text classification tasks are competitive with traditional models \citep{2, 4, 5}. 

The primary aim of this work is to develop a method to semantically classify columns in tabular data using character-level convolutional neural networks as a building block of a larger Automatic Machine Learning (AutoML) system. The choice of employing a character-level model, as opposed to a word-level or sentence-level one, enables misspellings and other social media language features, e.g., emoticons, to be naturally learned \citep{2}, while minimizing feature engineering and eliminating the constraint of being specialized to a particular language or style. Once classified into some basic classes, e.g., integer, string, categorical, geographic location, float, the columns can be routed to appropriate analyses for their assigned class by the AutoML system, e.g., categorical columns can be used in classification tasks, float columns in regression tasks, text columns can be summarized and physical locations can be plotted on a map. The particular AutoML system we are motivated by is being developed as part of DARPA's Data Driven Discovery of Models (D3M) program, and has been previously described, for e.g., in \cite{41}. We emphasize that the choice of tabular datasets is more general than might initially appear, and we will demonstrate in the numerical examples how other text classification problems may be cast in this form simply by organizing the relevant data into appropriate tabular form.

Our method first learns an initial set of weights from simulated data for some basic or base classes, and then employs transfer learning on hand-labeled data from public datasets identified and collected through CKAN to improve them to account for realistic data imperfections. Additionally, the transfer learning step is able to expand the set of handled classes from the base class set with reduced labeled data and computing power requirements.
 
A secondary purpose of this work is to develop an open source text classification software framework that can be applied to the broader text classification problem. We present an open source python library SIMON, an acronym for Semantic Inference for the Modeling of ONtologies, which aims to achieve this goal in a user-friendly and scalable/parallelizable fashion. To illustrate the effectiveness and versatility of the approach, we present experimental results on semantic classification of tabular data, age group prediction based on social media language, and classification of email spam.

To our knowledge, this is the first application of character-level CNNs to the problem of data type or \emph{semantic classification of tabular data}. 

\section{Background}

{\bf Text Classification}

Until recently, most successful approaches for text classification have resorted to tokenizing a string of characters and then applying some simple matching word statistics \citep{1}. These techniques perform very well for a defined domain, but require a predefined dictionary of words to overcome the lack of contextual information and extensive feature engineering to handle specific style variations \citep{2, 7, 8}. 

These limitations have motivated research on classifying text using more advanced machine learning techniques. A notable class of such methods is probabilistic latent topic models, which learn to map word occurrences to topics extracted from domain-related datasets. A traditional approach such as Support Vector Machines (SVM), k-Nearest Neighbor (kNN), or Naive Bayes is employed to achieve classification \citep{9, 10}. Another well-known approach is to use word2vec to generate features for text. This combines two main learning algorithms: continuous bag-of-words and continuous skip-gram, which introduces the extra semantic features that aid in text classification \citep{11, 12, 25}. These and similar models still employ a background knowledge repository to create an engineered language model \citep{1, 13}. 

Features generated using word2vec or a lookup table which are then fed into word-level convolutional neural network (CNN) yield competitive results to traditional text classification models \citep{4, 5, 14, 27, 30, 31}. Character-level features for classifying text have also been explored in the literature, specifically as character-level n-grams combined with linear classifiers or by incorporating character-level features to form a distributed representation for CNNs \citep{7, 8, 23, 24, 26}. It has recently been shown that applying convolutional neural networks only on characters can achieve competitive results when trained on large datasets, without the knowledge of words and not requiring any prior knowledge about the semantic and syntactic structure of a language \citep{2}. 

\newpage
{\bf Transfer Learning}

Transfer learning is another critical method in machine learning practice. The method is particularly useful in cases when labeled data is difficult to attain or is limited, because a previously trained model in a similar domain, that does not have data limitations, can be used \citep{36}. Transfer learning can even still effective when there is no class alignment between the trained model and the new target data. Recently transfer learning has been demonstrated to be effective between semantically related categories, as well as in cases where they are unrelated \citep{37, 38}. Fine-tuning neural networks on the target domain, which have been pre-trained on a source domain, is an idea that originated in computer vision, remaining relatively under-explored in natural language processing (NLP) tasks \citep{36}. Transfer learning in NLP has previously been applied to sentence-pair classification, slot tagging, and entity recognition \citep{36, 39, 40}.

\section{Methodology}
\label{headings}

In this section, the methodology behind the semantic classification method is formally presented. 

\subsection{Multi-Label Classification Problem Formulation}

Let $\mathcal{X}$ be the input space and $\mathcal{Y}$ be the output space, with $\mathcal{D}$ denoting an unknown distribution according to which the input points are drawn. We are interested in a multi-label setting, with $\mathcal{Y}=\{0,1\}^{n_{labels}}$ where $n_{labels}$ is the number of labels, several of which can be assigned to any given input example column. For instance, a column containing a city name on every row --- Austin, New York City, San Francisco, etc. --- may be labelled as both \emph{text} {\bf and} \emph{city}. Positive components of a vector in $\mathcal{Y}=\{0,1\}^{n_{labels}}$ indicate the classes associated with the input example column \citep{15}.

Given a set of labeled data $$S=\left( ({\bf x}^1,{\bf y}^1),...,({\bf x}^m,{\bf y}^m)\right)\in \left( \mathcal{X}\times\mathcal{Y} \right)^m,$$ with the training example vectors ${\bf x}^1,...,{\bf x}^m$ drawn i.i.d. according to $\mathcal{D}$, and $f:\mathcal{X}\rightarrow\mathcal{Y}$ with $y^i=f(x^i),\forall i\in\mathcal{I}\equiv\{i\}_{i=1}^{m}$ being the target labeling function, our goal is to find a hypothesis $h\in\mathcal{H}$ so as to minimize w.r.t $f$ the generalization error $$ \mathbb{E}_{x^i\sim\mathcal{D}}[1_{h(x^i)\neq f(x^i)}].$$ Here, $1_{ev}$ is the \emph{indicator function} of the event $ev$. 

\subsection{Neural Net Architecture}

This section introduces the architectural design of SIMON's neural network, which is trained on the labeled dataset to generate $h$. The design is modular, consisting of multiple convolutional layers and multiple bi-directional Long Short-Term Memory (LSTM) layers. First, the theory behind each of these layer types is described in the following subsections. The last subsection aggregates these layer types into the overall architecture. 

\subsubsection{Convolutional Layer}

Each convolutional layer computes one-dimensional convolutions between a vector of \emph{input features} and a 
number of different \emph{weight vectors}, called \emph{kernel functions} or \emph{filters}. For example, let ${\bf x} \in \mathbb{R}^{n_{\bf x}}$ be the vector of input features, where each ${\bf x}_i$ is a character from the input sentence. Meanwhile, let ${\bf w} \in \mathbb{R}^{n_{\bf w}}$ be one of the weight vectors. The convolution ${\bf c}$, between ${\bf x}$ and ${\bf w}$, can be thought of as the dot product between ${\bf w}$ and each sub-sequence of length $n_{\bf w}$ in ${\bf x}$, i.e., 

$$ {\bf c}_i = {\bf w}^T {\bf x}_{(i-n_{\bf w}+1):i}.$$ 

Specifically, ${\bf x}_{(i-n_{\bf w}+1):i}$ is a sub-sequence of ${\bf x}$ that ranges from indices $i-n_{\bf w}+1$ to $i$. Additionally, we define the convolution narrowly, which means that the dot product is only defined on sub-sequences of length $n_{\bf w}$ that fully overlap with ${\bf x}$. Therefore, $i$ ranges from $n_{\bf w}$ to $n_{\bf x}$. Furthermore, we set the stride length of the convolution to 1, which means that we advance one unit forward in ${\bf x}$ for each application of the dot product. Thus, the length of the output convolution ${\bf c}$ is $n_{\bf c}=n_{\bf x} - n_{\bf w} + 1$. Additionally, each convolutional layer contains several output convolutions, corresponding to the number of different weight vectors in that convolutional layer \citep{2,16}.

After applying the convolutions, we introduce non-linearity through a non-linear \emph{activation function}. Specifically, we use the rectified linear unit (ReLU) function $f(x) = \max\{0,x\}$. Finally, we initialize the weight vectors using the Glorot normal initializer, which draws samples from a normal distribution with mean $0$ and standard deviation $\sqrt{\frac{2}{n_{\bf x} + n_{\bf c}}}$. 

\subsubsection{Dropout Layer}

Each dropout layer removes a random set of input units by setting a fraction of these input units to zero during each update step of the training process. This improves the performance of the model by reducing overfitting \citep{17}.

\subsubsection{Max-Pooling Layer}

Each max-pooling layer computes the maximum value over a certain defined length. This is also known as \emph{temporal} max-pooling and is the one-dimensional equivalent of \emph{spatial} max-pooling used in computer vision applications \citep{18}.

\subsubsection{Bidirectional LSTM}

Recurrent neural networks became quite popular in the 1990s because of their ability to operate on sequential information. However, RNNs suffered from the vanishing gradient problem, where the network's output decays or explodes exponentially as it cycles through the network's recurrent connections \citep{19}. One solution to this vanishing gradient problem is the LSTM architecture proposed by Hochreiter and Schmidhuber in 1997 \citep{20}.

Each node in the LSTM architecture is a complex unit called a \emph{memory cell}, whose contents is modulated by an \emph{input gate}, an \emph{output gate}, and a \emph{forget gate}. These three gates control what information is written to, read from, and deleted from the cell. Each of the three gates receive all of the current and past inputs to the cell and combine these inputs according to a unique set of weights. Using the \emph{logistic sigmoid function}, each gate then squashes the output of this combination to an activation value between $0$ and $1$. More specifically, the activation of the output gate at time $t$, $ y_{out}(t)$ is 

$$ y^{out}(t) = \frac{1}{1+e^{(\sum_{i} w^{out}_iy^{out}_{i}(t-1))}},$$

where $i$ ranges over all the current and past inputs. Similarly, $ y^{in}(t)$ defines the activation of the input gate at time $t$ and $y^{forget}(t)$ defines the activation of the forget gate at time $t$.

The current and past inputs to the cell also pass through another activation function, $g$, which is usually the \emph{tanh} function or the logistic sigmoid function. The difference between this function and the aforementioned input gate is that this function passes the actual input values, while the aforementioned input gate moderates which of these inputs should be written to the cell. Therefore, the internal state of the memory cell, $s(t)$, at time $t$ is 

$$ s(t) = y^{forget}(t)s(t-1) + y^{in}(t)g(\sum_{i} w^{in}_i y_i^{in}(t-1)).$$

Finally, this internal state gets squashed by another activation function, $a$, also usually a tanh or logistic sigmoid function, and multiplied by the output gate to become the output of the memory cell. Thus, the output of the memory cell $o(t)$, at time $t$ is

$$ o(t) = y^{out}(t)a(s(t)).$$

The LSTM layers in our model are also bidirectional, which means that the input sequence is fed through two separate LSTMs, once forward and once backward. Both LSTMs then connect to the same output layer. This allows the output layer to access information from both the previous values of the sequence and the future values of the sequence at each time step $t$ \citep{19, 20}.

\subsubsection{Overall SIMON architecture}

SIMON's architecture consists of two components---one that encodes each individual sentence and one that subsequently encodes the document as a whole. In the context of tabular data, ``document'' corresponds to a column and ``sentence`` corresponds to a cell within that column. The architecture is influenced heavily by \cite{2}. 

The network that encodes each individual sentence connects thirteen layers. Its input is the sentence's sequence of characters shortened to a maximum length of $max_{len}$ characters (20 in the case of semantic classification of tabular data, tunable and different for the other applications, as will be reported accordingly). This input is connected to a one-hot encoding layer that contains a dictionary of 71 characters. This includes 26 English letters, 10 digits, new line and 33 other characters [6]. One-hot encoding in a layer, as opposed to adding a dimension to the input data directly, provides significant memory advantages.

The architecture continues with convolutional layer with a kernel length of one and a kernel dimension of 40. This is connected to a dropout layer with a dropout probability of 0.1 and a max-pooling layer with a pool size of 2. Another convolutional layer follows, this time with a kernel length of three and a kernel dimension of 200. This convolutional layer is connected to another dropout layer with a dropout probability of 0.1 and another max-pooling layer with a pool size of 2. This pooling layer is then connected to a third convolutional layer, which again has a kernel length of three, but this time has a kernel dimension of 1000. This convolutional layer is then connected to another dropout layer that has a dropout probability of 0.1 and another max-pooling layer that has a pool size of 2. The output of this pooling layer is a one-dimensional vector with 1000 features. Next, this output vector is connected to a bidirectional LSTM that contains 256 units. 20\% of the inputs to the input gates and 20\% of the recurrent connections are dropped. Finally, the two LSTM directions are merged into a layer with 512 units and fed through a dropout layer with a dropout probability of 0.3. The final sentence encoding is the output of this dropout layer. 

On the other hand, the network that encodes the document as a whole connects seven layers. The first step takes the first $max_{cells}$ rows (500 in the case of semantic classification, tunable and different for other applications) of the document and feeds each of these sentences through the sentence encoder. We note that when a raw input column is shorter than $max_{cells}$, uniformly sampled cells are appended to make its length equal to $max_{cells}$ exactly. Similarly, when the column is longer, uniformly randomly sampled cells are removed. The result is a document of length $max_{cells}$ that contains 512 features per sentence. This input is connected to a bidirectional LSTM with 128 units and then merged into a layer that contains 256 units. Similarly to the sentence encoder, 20\% of the inputs to the input gates and 20\% of the recurrent connections are dropped. Next, this output is fed through a dropout layer with dropout probability 0.3 before connecting to a dense layer with 128 units and ReLU activations. Finally, this densely connected layer passes through a dropout layer with dropout probability 0.3 and then connects to another densely connected layer, this time containing a number of units equal to the number of categories and sigmoid activation functions. The output of this final layer contains the probabilities of classifying the document as each different type of class. Probability above threshold $p_{threshold}$ (typically $0.5$, or tuned from there) results in the corresponding input vector being classified as the corresponding class.

\subsection{Transfer Learning}

Synthetic data was initially generated for some basic, or base, semantic classes using the python library Faker \footnote{\url{https://github.com/stympy/faker}}. The base semantic classes are \emph{address, boolean, datetime, email, float, int, phone, text, uri}. Approximately 10,000 such synthetic columns were generated and used to produce an initial set of weights. 

The final fully-connected layer of the network was then stripped away, and the initial network frozen and treated as a feature-extractor. Real open-source data columns were collected from CKAN and labeled in a semi-supervised fashion, as described in the next subsection, to yield approximately 10,000 real labeled data columns covering the base classes. Furthermore, simple statistical rules/heuristics were used to label these columns as additionally \emph{categorical} and \emph{ordinal}, for e.g., small fraction of a column being occupied by unique values implies categorical, categorical and numerical, i.e., int or float, implied ordinal. Thus, the number of handled classes was increased by two. A re-sized final fully-connected layer was then trained on top of the aforementioned initial network frozen and treated as a feature-extractor. In so doing, the network learned the imperfections inherent in real data, and was also able to expand the set of classes handled. The improvement in accuracy from learning real data imperfections is apparent in Figure \ref{fig1}, where the validation binary accuracy can be seen to increase from the initial $\approx87.5\%$ to greater than 98\% during the transfer learning experiment.  

In another variant of the transfer learning experiment, the real data was further augmented with columns of geographical data from the open-source \emph{GeoNames} repository \footnote{\url{https://www.geonames.org}}, covering the additional classes \emph{state, city, postal code, latitude, longitude, country} and \emph{country code}. The number of handled classes thereby grew further by 7, to yield 18 classes total. The number of augmenting geographical data columns generated was 1000 per additional geo class, yielding 7000 additional labeled data columns.

\subsection{Data Labeling Procedure}

In order to produce training data for the transfer learning step, we first used a simple model trained on Faker data to predict labels on real data. For each data type, columns were generated, and for each such column we calculated the average number each of decimals, letters, and punctuation marks to use as input features. For each data type, a single Random Forest classifier was trained in a one-vs-all fashion on that data type, reaching an accuracy of \textgreater 90\% on the Faker data. Next, real data was collected from public data sources identified on CKAN \footnote{\url{https://ckan.org/about/instances/}} and collected using the CKAN api \footnote{\url{https://docs.ckan.org/en/ckan-2.7.0/api/}}. The aforementioned classifiers were then used to predict labels for this more realistic data, the results of which were first exhaustively manually checked for correctness, corrected if needed and then used as training data for the SIMON semantic tabular data classifier. The training data thereby collected is a collection of approximately 10,000 labeled real-world data columns spanning the base class set.

\section{Numerical Results}
\label{headings_2}

In this section, training convergence and performance results from the SIMON classifier are presented for a diverse set of text classification tasks. Extensive testing results are first presented for the semantic classification problem in the first two examples. The following two examples demonstrate the flexibility of the SIMON framework and software infrastructure via applicability to two related but different problems - age classification from twitter text, and email spam classification.    

\subsection{Training}

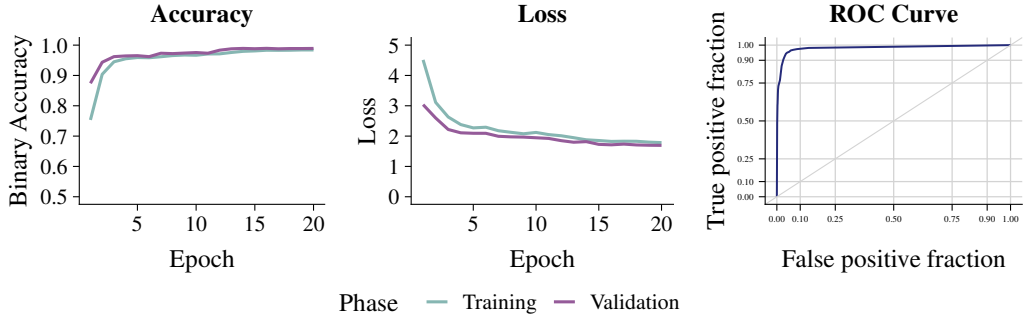
\begin{figure}[h]
  \centering
    \makebox[\textwidth][c]{\resizebox{1.00\textwidth}{!}{\input{figs/figure_1}}}
  \caption{Convergence and performance results from the tabular data classification task.}
  \label{fig1}
\end{figure}

Figure \ref{fig1} shows the convergence of the transfer learning experiment with the geographical categories included (a total of 18 handled categories). \emph{Binary accuracy}---the ratio of the number of correct predictions, i.e., true positives and true negatives relative to the total number of samples---is the metric shown.

Data was segmented according to the standard 60\% train, 30\% validate and 10\% test split. Upon convergence, training binary accuracy of 98.4\%, validation binary accuracy of 98.9\% and test binary accuracy of 98.8\% are attained. Training took 1.59 hours on an NC6 VM size on Azure, which possesses one NVIDIA Tesla K80 GPU. By comparison, initial training on faker data took approximately 6 hours on the same machine. Precision at threshold probability of 0.5 was computed as 0.87, recall as 0.81 and F1 score as 0.84. The ROC curve is also shown in the Figure.

\subsection{Evaluation}

Extensive testing results are first presented for the semantic classification problem in the first two examples. The following two examples demonstrate the flexibility of the SIMON framework via applicability to two different problems - age classification from twitter text, and email spam classification.

\subsubsection{Tabular Semantic Classification Evaluation on D3M Datasets}

In Table 1 of the Appendix, we present a {\it SIMON similarity score} that compares SIMON's annotations to manually annotated D3M datasets which are provided by DARPA's D3M program \footnote{\url{https://gitlab.com/datadrivendiscovery}}. The SIMON similarity score is defined as the percentage of labels for which any of SIMON's annotations matches the manual annotation. Note that SIMON is a multi-class multi-label classifier that can produce several categories for each tabular column, but for the purposes of this comparison, it is restricted by the single-label manual annotation. 

With geographical categories included, SIMON performs well with an average similarity score of {\bf 0.92} across 38 datasets. Additionally, the tool outperforms the manual annotations with geographic category detection - notably in cases where classifications are longitude, latitude, postal code, and country. For example, for the dataset \texttt{26\textunderscore radon\textunderscore seed}, SIMON correctly matches the category ``zip'' to ``postal code'', whereas that column was manually annotated as just ``categorical''. Such cases lead to a lower reported similarity score, even though these additional annotations provide a better understanding of the dataset beyond the manual annotation. Other interesting cases include \texttt{299\textunderscore librasmove}, where SIMON’s annotations correctly match all 92 manual annotations, and \texttt{1491\textunderscore one\textunderscore hundred\textunderscore plants\textunderscore margin}, where SIMON’s annotations correctly match all 66 manual annotations for both datasets. 

Manual annotations were also compared to automatic \texttt{pandas.DataFrame.dtypes()} annotations, with the assumption that \texttt{pandas's} \texttt{object} classification encompasses the data type string. Given that \texttt{pandas} and the manual annotations both have single labels, the similarity score is defined as the number of matching annotations divided by the total number of annotations. These results achieved an average similarity score of {\bf 0.71}, which demonstrates that SIMON’s annotations matched the manual annotations more closely than \texttt{pandas}. For a good majority of the differences, \texttt{pandas} classified columns as an \texttt{object}, where SIMON’s annotations were more literal. For example, for the dataset \texttt{LL0\textunderscore 1100\textunderscore popularkids}, there were five columns that were classified as an \texttt{object} according to \texttt{pandas}, which SIMON labeled as categorical. An example column is \texttt{gender}, where the only two possible values are \texttt{boy} and \texttt{girl}.

\subsubsection{Tabular Semantic Classification Evaluation on Paleontology data.world Dataset}

Evaluation of performance on this dataset highlights potential limitations of the SIMON classifier in its current form, and provides an opportunity to discuss issues that arise when dataset column truncation comes into play. This dataset contains 112 columns, some of which are longer than 900,000 rows. It contains information about some dinosaur fossils, where they were discovered and other pertinent information, curated by The Palebiology Database. It is freely available as dataset \emph{paleobiodb} on data.world.

Binary testing accuracy achieved was 97.7\%. Lowest class binary accuracy was 87.5\% for the categorical class, likely due to the fact that the very large columns needed to be subsampled to fit the fixed-length neural network expected size of 500 rows, and thereby lost some of the information needed to correctly classify those columns as categoricals. For this reason, in the software implementation of the method we added the ability for the user to replace neural net predictions for this class, with those based on simple statistical heuristics calculated on the entire column (if neural net predictions can be reasonably anticipated to be incorrect, as in this case). This is significantly more expensive computationally for very large columns, but does yield near-perfect accuracy for the categorical/ordinal classes. Additionally, fossil feature names tend to bear similarity to physical locations of their discovery, which led to an observable increase in false positive rates for the geographical categories in this example. 

\subsubsection{Twitter Age Prediction}

Prediction of user age group based on tweets was performed. Twitter data publicly self-identified to fall into one of three age groups by the user --- 14 to 17 years, 18 to 23 years, 24 plus years --- was collected. This collection results in a corpus of 10,000 total tweets. Tweets were arranged into ten-cell columns by category, i.e., $max_{cells}=10$, and each cell was truncated at $max_{len}=280$ characters to accommodate the whole tweet length. Thereby, the age classification problem was cast as a tabular semantic classification problem, and the same network architecture was trained on the twitter data.

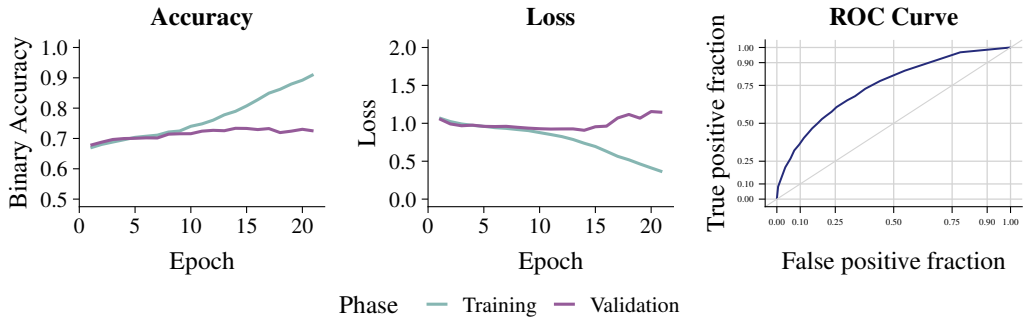
\begin{figure}[h]
  \centering
  \makebox[\textwidth][c]{\resizebox{1.00\textwidth}{!}{\input{figs/figure_2}}}
  \caption{Training convergence information for the twitter age prediction experiment.}
  \label{fig2}
\end{figure}

Figure \ref{fig2} presents the training results for the age prediction task. Data was segmented according to the standard 60\% train, 30\% validate and 10\% test split. Upon convergence, training \emph{binary} accuracy of 79.0\%, validation binary accuracy of 73.4\% and test binary accuracy of 70.9\% were attained. Training was performed from scratch (without any transfer learning), and took 35.6 minutes to reach the best checkpoint (number 14) on the aforementioned NC6 VM size on Azure, after which over-fitting can be clearly observed. Corresponding precision, at threshold probability of 0.5, was calculated as 0.57, recall as 0.54, and F1 score as 0.55. The ROC curve is also shown in the Figure.

While these results appeared disappointing initially, comparing this accuracy to uniform random chance --- which corresponds to recall and precision of 0.33 --- elucidates the potential usefulness of the result. We note that state-of-the-art results for age prediction, which exploit network structure and other user metadata, are comparable. For instance, \cite{42} reports a \emph{correlation coefficient} of 0.77. Our results could likely be improved further by introducing network features, and optimizing other hyper-parameters. The ability of SIMON to quickly predict the age group of users from tweet text NLP {\bf only}, relatively well, is noteworthy.

\subsubsection{Email Spam Classification}

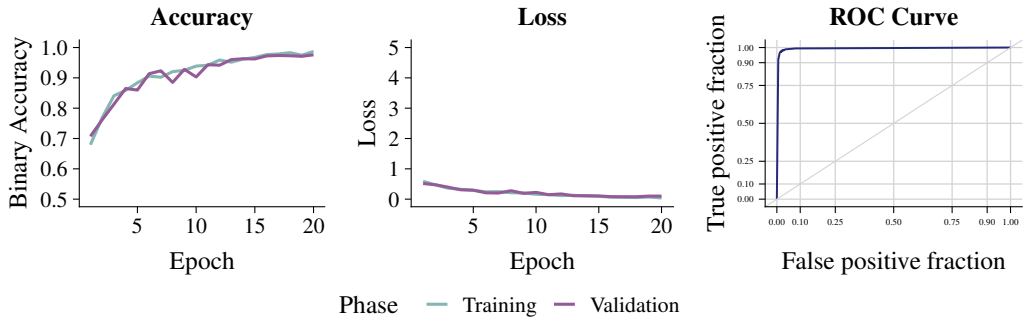
\begin{figure}[h]
  \centering
    \makebox[\textwidth][c]{\resizebox{1.00\textwidth}{!}{\input{figs/figure_3}}}
  \caption{Convergence and performance results for the email spam classification task.}
  \label{fig3}
\end{figure}

Emails were parsed into columns, with one sentence per cell, and up to a maximum of $max_{cells}=500$ sentences/cells. Each sentence/cell was truncated at $max_{len}=100$ characters.  The popular Enron email dataset \footnote{ \url{www.kaggle.com/wcukierski/enron-email-dataset}} was labeled as \emph{ham}, and a 419 spam fraud corpus \footnote{ \url{www.kaggle.com/rtatman/fraudulent-email-corpus}} was labeled as \emph{spam}. In this fashion, a tabular dataset of 10000 emails/columns balanced between the two classes was generated, and the SIMON algorithmic software was trained on it. To date, the best convergence result is 98.7\% training, 97.6\% validation and 97.9\% testing accuracies (see Figure \ref{fig3}, the ROC curve is also shown). Precision, F1 score and recall were all computed during the testing phase as 97.9\%. Only a small fraction of the Enron dataset ($\approx$2\%) was used in training, and no hyper-parameter (e.g. $max_{cells}$ and $max_{len}$) tuning was performed beyond initial choices. This leads us to believe that these metrics could readily be improved further (an ongoing effort by the authors), and argue that this performance is already notable. It is within 1-2\% of state-of-the-art accuracies of 99+\%, associated with approaches that employ email headers in the analysis --- whereas SIMON employs email text NLP {\bf only}.

\section{Discussion}
Semantic classification of tabular data achieved excellent performance. Surprisingly to these authors, the CNN was able to classify with high accuracy categories based on column-level statistical properties, i.e., categorical and ordinal. Our successful transfer learning experience reported here adds more evidence for its potential in NLP, evidence which has been relatively lacking in this domain \citep{36}. The ability of the framework to predict user age solely from social media natural language on twitter (without using network information or any other user metadata) is noteworthy. The spam classification example also demonstrates good performance in initial experiments, close to the performance of state-of-the-art spam classifiers which additionally employ header information and other metadata. These observations suggest that these very important problems may be competitively solved by analyzing the semantic structure of natural language alone at the character level. The flexibility and power of the SIMON text-classification framework is clearly demonstrated by these examples.  

\section{Software Implementation - SIMON}

A software implementation of SIMON\footnote{Located, with instructions, at \url{https://github.com/NewKnowledge/simon}} was developed, able to scale to multi-GPU settings for larger problems. Extensions leveraging the Horovod python library for full heterogeneous multi-GPU server scaling are in development. It is a pip-installable python library running on top of Keras, developed as part of the authors' participation in DARPA's D3M program.

\subsubsection*{Acknowledgments}

Work was supported by the Defense Advanced Research Projects Agency (DARPA) under Contract Number D3M (FA8750-17-C-0094). This work was also supported, in part, by the Defense Advanced Research Projects Agency (DARPA) (contract number ASED (FA8650-18-C-7889)). Views, opinions, and findings contained in this report are those of the authors and should not be construed as an official Department of Defense position, policy, or decision.

\nocite{28}
\bibliographystyle{newapa}

\newpage
\section*{Appendix -- Evaluation Results on D3M Datasets}

\begin{table}[!htbp]
    \centering
    \begin{tabular}{|p{4cm}||P{2.5cm}|P{2.5cm}|P{2.5cm}|}
    \hline
    \centering
    \textbf{Dataset} & \textbf{Matching Annotations} & \textbf{Unique Annotations} & \textbf{Similarity Score} \\
    \hline
     1. 22\_handgeometry & 3 & 0 & 1.00 \\
     2. 26\_radon\_seed & 15 & 15 & 0.50 \\
     3. 27\_WordLevels & 13 & 1 & 0.93 \\
     4. 30\_personae & 7 & 0 & 1.00 \\
     5. 31\_urbansound & 5 & 1 & 0.83 \\
     6. 32\_wikiqa & 4 & 0 & 1.00 \\
     7. 38\_sick & 29 & 2 & 0.94 \\
     8. 49\_facebook & 4 & 0 & 1.00 \\
     9. 56\_sunspots & 4 & 0 & 1.00 \\
     10. 59\_umls & 5 & 0 & 1.00 \\
     11. 60\_jester & 4 & 0 & 1.00 \\
     12. 66\_chlConcentration & 3 & 0 & 1.00 \\
     13. 185\_baseball & 18 & 1 & 0.95 \\
     14. 196\_autoMpg & 9 & 0 & 1.00 \\
     15. 299\_librasmove & 92 & 0 & 1.00 \\
     16. 313\_spectrometer & 102 & 2 & 0.98 \\
     17. 534\_cps\_85\_wages & 12 & 0 & 1.00 \\
     18. 1491\_100\_plants & 66 & 0 & 1.00 \\
     19. 1567\_poker\_hand & 12 & 0 & 1.00 \\
     20. 4550\_MiceProtein & 82 & 1 & 0.99 \\
     21. 6\_70\_com\_amazon & 3 & 0 & 1.00 \\
     22. 6\_86\_com\_DBLP & 3 & 0 & 1.00 \\
     23. DS01876 & 3 & 0 & 1.00 \\
     24. LL0\_1100\_popularkids & 11 & 1 & 0.92 \\
     25. LL0\_186\_braziltourism & 9 & 1 & 0.90 \\
     26. LL0\_207\_autoPrice & 17 & 0 & 1.00 \\
     27. LL0\_acled & 29 & 2 & 0.94 \\
     28. LL0\_acled\_reduced & 17 & 11 & 0.61 \\
     29. LL1\_336\_MS\_Geolife & 5 & 3 & 0.63 \\
     30. 336\_MS\_Geolife\_2 & 7 & 2 & 0.78 \\
     31. LL1\_736\_stock\_market & 6 & 0 & 1.00 \\
     32. LL1\_pedestrian & 2 & 1 & 0.67 \\
     33. uu1\_datasmash & 3 & 0 & 1.00 \\
     34. uu2\_gp\_hyperparameter & 4 & 0 & 1.00 \\
     35. uu2\_gp\_hp\_v2 & 4 & 0 & 1.00 \\
     36. uu3\_world\_dev\_ind & 6 & 1 & 0.86 \\
     37. uu4\_SPECT & 46 & 22 & 0.68 \\
     38. 57\_hypothyroid & 29 & 2 & 0.94 \\
     
     \hline
    \end{tabular}
\caption{Comparison of Manual D3M Annotations to Annotations.}
\label{Comparison of Manual D3M Annotations to SIMON Annotations}
\end{table}
\FloatBarrier

\end{document}

%% file: figs/figure_1.tex
\begin{tikzpicture}[x=1pt,y=1pt]
\definecolor{fillColor}{RGB}{255,255,255}
\path[use as bounding box,fill=fillColor,fill opacity=0.00] (0,0) rectangle (578.16,180.67);
\begin{scope}
\path[clip] ( 50.12, 62.13) rectangle (185.72,154.36);
\definecolor{drawColor}{RGB}{134,182,178}

\path[draw=drawColor,line width= 1.7pt,line join=round] ( 56.28,108.80) --
	( 62.77,134.02) --
	( 69.26,140.95) --
	( 75.75,142.62) --
	( 82.24,143.31) --
	( 88.72,143.19) --
	( 95.21,143.75) --
	(101.70,144.39) --
	(108.19,144.69) --
	(114.68,144.57) --
	(121.16,145.27) --
	(127.65,145.34) --
	(134.14,146.10) --
	(140.63,146.73) --
	(147.12,146.97) --
	(153.60,147.34) --
	(160.09,147.30) --
	(166.58,147.34) --
	(173.07,147.50) --
	(179.56,147.50);
\definecolor{drawColor}{RGB}{149,91,153}

\path[draw=drawColor,line width= 1.7pt,line join=round] ( 56.28,129.06) --
	( 62.77,140.70) --
	( 69.26,143.80) --
	( 75.75,144.20) --
	( 82.24,144.34) --
	( 88.72,143.80) --
	( 95.21,145.69) --
	(101.70,145.48) --
	(108.19,145.78) --
	(114.68,146.06) --
	(121.16,145.69) --
	(127.65,147.37) --
	(134.14,148.13) --
	(140.63,148.36) --
	(147.12,148.18) --
	(153.60,148.39) --
	(160.09,148.14) --
	(166.58,148.28) --
	(173.07,148.28) --
	(179.56,148.34);
\end{scope}
\begin{scope}
\path[clip] (  0.00,  0.00) rectangle (578.16,180.67);
\definecolor{drawColor}{RGB}{0,0,0}

\path[draw=drawColor,line width= 0.6pt,line join=round,line cap=rect] ( 50.12, 62.13) --
	( 50.12,154.36);
\end{scope}
\begin{scope}
\path[clip] (  0.00,  0.00) rectangle (578.16,180.67);
\definecolor{drawColor}{RGB}{0,0,0}

\node[text=drawColor,anchor=base east,inner sep=0pt, outer sep=0pt, scale=  1.20] at ( 43.62, 62.19) {0.5};

\node[text=drawColor,anchor=base east,inner sep=0pt, outer sep=0pt, scale=  1.20] at ( 43.62, 78.96) {0.6};

\node[text=drawColor,anchor=base east,inner sep=0pt, outer sep=0pt, scale=  1.20] at ( 43.62, 95.73) {0.7};

\node[text=drawColor,anchor=base east,inner sep=0pt, outer sep=0pt, scale=  1.20] at ( 43.62,112.50) {0.8};

\node[text=drawColor,anchor=base east,inner sep=0pt, outer sep=0pt, scale=  1.20] at ( 43.62,129.27) {0.9};

\node[text=drawColor,anchor=base east,inner sep=0pt, outer sep=0pt, scale=  1.20] at ( 43.62,146.04) {1.0};
\end{scope}
\begin{scope}
\path[clip] (  0.00,  0.00) rectangle (578.16,180.67);
\definecolor{drawColor}{RGB}{0,0,0}

\path[draw=drawColor,line width= 0.6pt,line join=round] ( 46.62, 66.32) --
	( 50.12, 66.32);

\path[draw=drawColor,line width= 0.6pt,line join=round] ( 46.62, 83.09) --
	( 50.12, 83.09);

\path[draw=drawColor,line width= 0.6pt,line join=round] ( 46.62, 99.86) --
	( 50.12, 99.86);

\path[draw=drawColor,line width= 0.6pt,line join=round] ( 46.62,116.63) --
	( 50.12,116.63);

\path[draw=drawColor,line width= 0.6pt,line join=round] ( 46.62,133.40) --
	( 50.12,133.40);

\path[draw=drawColor,line width= 0.6pt,line join=round] ( 46.62,150.17) --
	( 50.12,150.17);
\end{scope}
\begin{scope}
\path[clip] (  0.00,  0.00) rectangle (578.16,180.67);
\definecolor{drawColor}{RGB}{0,0,0}

\path[draw=drawColor,line width= 0.6pt,line join=round,line cap=rect] ( 50.12, 62.13) --
	(185.72, 62.13);
\end{scope}
\begin{scope}
\path[clip] (  0.00,  0.00) rectangle (578.16,180.67);
\definecolor{drawColor}{RGB}{0,0,0}

\path[draw=drawColor,line width= 0.6pt,line join=round] ( 82.24, 58.63) --
	( 82.24, 62.13);

\path[draw=drawColor,line width= 0.6pt,line join=round] (114.68, 58.63) --
	(114.68, 62.13);

\path[draw=drawColor,line width= 0.6pt,line join=round] (147.12, 58.63) --
	(147.12, 62.13);

\path[draw=drawColor,line width= 0.6pt,line join=round] (179.56, 58.63) --
	(179.56, 62.13);
\end{scope}
\begin{scope}
\path[clip] (  0.00,  0.00) rectangle (578.16,180.67);
\definecolor{drawColor}{RGB}{0,0,0}

\node[text=drawColor,anchor=base,inner sep=0pt, outer sep=0pt, scale=  1.20] at ( 82.24, 47.37) {5};

\node[text=drawColor,anchor=base,inner sep=0pt, outer sep=0pt, scale=  1.20] at (114.68, 47.37) {10};

\node[text=drawColor,anchor=base,inner sep=0pt, outer sep=0pt, scale=  1.20] at (147.12, 47.37) {15};

\node[text=drawColor,anchor=base,inner sep=0pt, outer sep=0pt, scale=  1.20] at (179.56, 47.37) {20};
\end{scope}
\begin{scope}
\path[clip] (  0.00,  0.00) rectangle (578.16,180.67);
\definecolor{drawColor}{RGB}{0,0,0}

\node[text=drawColor,anchor=base,inner sep=0pt, outer sep=0pt, scale=  1.40] at (117.92, 27.75) {Epoch};
\end{scope}
\begin{scope}
\path[clip] (  0.00,  0.00) rectangle (578.16,180.67);
\definecolor{drawColor}{RGB}{0,0,0}

\node[text=drawColor,rotate= 90.00,anchor=base,inner sep=0pt, outer sep=0pt, scale=  1.40] at ( 20.97,108.25) {Binary Accuracy};
\end{scope}
\begin{scope}
\path[clip] (  0.00,  0.00) rectangle (578.16,180.67);
\definecolor{drawColor}{RGB}{0,0,0}

\node[text=drawColor,anchor=base,inner sep=0pt, outer sep=0pt, scale=  1.40] at (117.92,162.69) {\bfseries Accuracy};
\end{scope}
\begin{scope}
\path[clip] (233.51, 62.13) rectangle (378.44,154.36);
\definecolor{drawColor}{RGB}{134,182,178}

\path[draw=drawColor,line width= 1.7pt,line join=round] (240.10,142.00) --
	(247.03,118.52) --
	(253.97,110.43) --
	(260.90,106.24) --
	(267.83,104.40) --
	(274.77,104.77) --
	(281.70,102.83) --
	(288.64,101.96) --
	(295.57,101.15) --
	(302.51,101.89) --
	(309.44,100.70) --
	(316.38,100.05) --
	(323.31, 98.98) --
	(330.25, 97.82) --
	(337.18, 97.41) --
	(344.11, 96.92) --
	(351.05, 97.03) --
	(357.98, 97.00) --
	(364.92, 96.52) --
	(371.85, 96.26);
\definecolor{drawColor}{RGB}{149,91,153}

\path[draw=drawColor,line width= 1.7pt,line join=round] (240.10,117.26) --
	(247.03,109.88) --
	(253.97,103.52) --
	(260.90,101.68) --
	(267.83,101.42) --
	(274.77,101.41) --
	(281.70, 99.76) --
	(288.64, 99.44) --
	(295.57, 99.31) --
	(302.51, 98.94) --
	(309.44, 98.58) --
	(316.38, 97.35) --
	(323.31, 96.47) --
	(330.25, 96.79) --
	(337.18, 95.29) --
	(344.11, 95.04) --
	(351.05, 95.43) --
	(357.98, 94.97) --
	(364.92, 94.80) --
	(371.85, 94.73);
\end{scope}
\begin{scope}
\path[clip] (  0.00,  0.00) rectangle (578.16,180.67);
\definecolor{drawColor}{RGB}{0,0,0}

\path[draw=drawColor,line width= 0.6pt,line join=round,line cap=rect] (233.51, 62.13) --
	(233.51,154.36);
\end{scope}
\begin{scope}
\path[clip] (  0.00,  0.00) rectangle (578.16,180.67);
\definecolor{drawColor}{RGB}{0,0,0}

\node[text=drawColor,anchor=base east,inner sep=0pt, outer sep=0pt, scale=  1.20] at (227.01, 62.19) {0};

\node[text=drawColor,anchor=base east,inner sep=0pt, outer sep=0pt, scale=  1.20] at (227.01, 78.96) {1};

\node[text=drawColor,anchor=base east,inner sep=0pt, outer sep=0pt, scale=  1.20] at (227.01, 95.73) {2};

\node[text=drawColor,anchor=base east,inner sep=0pt, outer sep=0pt, scale=  1.20] at (227.01,112.50) {3};

\node[text=drawColor,anchor=base east,inner sep=0pt, outer sep=0pt, scale=  1.20] at (227.01,129.27) {4};

\node[text=drawColor,anchor=base east,inner sep=0pt, outer sep=0pt, scale=  1.20] at (227.01,146.04) {5};
\end{scope}
\begin{scope}
\path[clip] (  0.00,  0.00) rectangle (578.16,180.67);
\definecolor{drawColor}{RGB}{0,0,0}

\path[draw=drawColor,line width= 0.6pt,line join=round] (230.01, 66.32) --
	(233.51, 66.32);

\path[draw=drawColor,line width= 0.6pt,line join=round] (230.01, 83.09) --
	(233.51, 83.09);

\path[draw=drawColor,line width= 0.6pt,line join=round] (230.01, 99.86) --
	(233.51, 99.86);

\path[draw=drawColor,line width= 0.6pt,line join=round] (230.01,116.63) --
	(233.51,116.63);

\path[draw=drawColor,line width= 0.6pt,line join=round] (230.01,133.40) --
	(233.51,133.40);

\path[draw=drawColor,line width= 0.6pt,line join=round] (230.01,150.17) --
	(233.51,150.17);
\end{scope}
\begin{scope}
\path[clip] (  0.00,  0.00) rectangle (578.16,180.67);
\definecolor{drawColor}{RGB}{0,0,0}

\path[draw=drawColor,line width= 0.6pt,line join=round,line cap=rect] (233.51, 62.13) --
	(378.44, 62.13);
\end{scope}
\begin{scope}
\path[clip] (  0.00,  0.00) rectangle (578.16,180.67);
\definecolor{drawColor}{RGB}{0,0,0}

\path[draw=drawColor,line width= 0.6pt,line join=round] (267.83, 58.63) --
	(267.83, 62.13);

\path[draw=drawColor,line width= 0.6pt,line join=round] (302.51, 58.63) --
	(302.51, 62.13);

\path[draw=drawColor,line width= 0.6pt,line join=round] (337.18, 58.63) --
	(337.18, 62.13);

\path[draw=drawColor,line width= 0.6pt,line join=round] (371.85, 58.63) --
	(371.85, 62.13);
\end{scope}
\begin{scope}
\path[clip] (  0.00,  0.00) rectangle (578.16,180.67);
\definecolor{drawColor}{RGB}{0,0,0}

\node[text=drawColor,anchor=base,inner sep=0pt, outer sep=0pt, scale=  1.20] at (267.83, 47.37) {5};

\node[text=drawColor,anchor=base,inner sep=0pt, outer sep=0pt, scale=  1.20] at (302.51, 47.37) {10};

\node[text=drawColor,anchor=base,inner sep=0pt, outer sep=0pt, scale=  1.20] at (337.18, 47.37) {15};

\node[text=drawColor,anchor=base,inner sep=0pt, outer sep=0pt, scale=  1.20] at (371.85, 47.37) {20};
\end{scope}
\begin{scope}
\path[clip] (  0.00,  0.00) rectangle (578.16,180.67);
\definecolor{drawColor}{RGB}{0,0,0}

\node[text=drawColor,anchor=base,inner sep=0pt, outer sep=0pt, scale=  1.40] at (305.97, 27.75) {Epoch};
\end{scope}
\begin{scope}
\path[clip] (  0.00,  0.00) rectangle (578.16,180.67);
\definecolor{drawColor}{RGB}{0,0,0}

\node[text=drawColor,rotate= 90.00,anchor=base,inner sep=0pt, outer sep=0pt, scale=  1.40] at (213.69,108.25) {Loss};
\end{scope}
\begin{scope}
\path[clip] (  0.00,  0.00) rectangle (578.16,180.67);
\definecolor{drawColor}{RGB}{0,0,0}

\node[text=drawColor,anchor=base,inner sep=0pt, outer sep=0pt, scale=  1.40] at (305.97,162.69) {\bfseries Loss};
\end{scope}
\begin{scope}
\path[clip] (429.12, 62.13) rectangle (571.16,154.36);
\definecolor{drawColor}{RGB}{255,255,255}

\path[draw=drawColor,line width= 0.7pt,line join=round] (429.12, 67.16) --
	(571.16, 67.16);

\path[draw=drawColor,line width= 0.7pt,line join=round] (429.12, 68.00) --
	(571.16, 68.00);

\path[draw=drawColor,line width= 0.7pt,line join=round] (429.12, 68.84) --
	(571.16, 68.84);

\path[draw=drawColor,line width= 0.7pt,line join=round] (429.12, 69.68) --
	(571.16, 69.68);

\path[draw=drawColor,line width= 0.7pt,line join=round] (429.12, 70.51) --
	(571.16, 70.51);

\path[draw=drawColor,line width= 0.7pt,line join=round] (429.12, 71.35) --
	(571.16, 71.35);

\path[draw=drawColor,line width= 0.7pt,line join=round] (429.12, 72.19) --
	(571.16, 72.19);

\path[draw=drawColor,line width= 0.7pt,line join=round] (429.12, 73.03) --
	(571.16, 73.03);

\path[draw=drawColor,line width= 0.7pt,line join=round] (429.12, 73.87) --
	(571.16, 73.87);

\path[draw=drawColor,line width= 0.7pt,line join=round] (429.12,142.62) --
	(571.16,142.62);

\path[draw=drawColor,line width= 0.7pt,line join=round] (429.12,143.46) --
	(571.16,143.46);

\path[draw=drawColor,line width= 0.7pt,line join=round] (429.12,144.30) --
	(571.16,144.30);

\path[draw=drawColor,line width= 0.7pt,line join=round] (429.12,145.14) --
	(571.16,145.14);

\path[draw=drawColor,line width= 0.7pt,line join=round] (429.12,145.98) --
	(571.16,145.98);

\path[draw=drawColor,line width= 0.7pt,line join=round] (429.12,146.82) --
	(571.16,146.82);

\path[draw=drawColor,line width= 0.7pt,line join=round] (429.12,147.66) --
	(571.16,147.66);

\path[draw=drawColor,line width= 0.7pt,line join=round] (429.12,148.49) --
	(571.16,148.49);

\path[draw=drawColor,line width= 0.7pt,line join=round] (429.12,149.33) --
	(571.16,149.33);

\path[draw=drawColor,line width= 0.7pt,line join=round] (436.86, 62.13) --
	(436.86,154.36);

\path[draw=drawColor,line width= 0.7pt,line join=round] (438.16, 62.13) --
	(438.16,154.36);

\path[draw=drawColor,line width= 0.7pt,line join=round] (439.45, 62.13) --
	(439.45,154.36);

\path[draw=drawColor,line width= 0.7pt,line join=round] (440.74, 62.13) --
	(440.74,154.36);

\path[draw=drawColor,line width= 0.7pt,line join=round] (442.03, 62.13) --
	(442.03,154.36);

\path[draw=drawColor,line width= 0.7pt,line join=round] (443.32, 62.13) --
	(443.32,154.36);

\path[draw=drawColor,line width= 0.7pt,line join=round] (444.61, 62.13) --
	(444.61,154.36);

\path[draw=drawColor,line width= 0.7pt,line join=round] (445.90, 62.13) --
	(445.90,154.36);

\path[draw=drawColor,line width= 0.7pt,line join=round] (447.19, 62.13) --
	(447.19,154.36);

\path[draw=drawColor,line width= 0.7pt,line join=round] (553.08, 62.13) --
	(553.08,154.36);

\path[draw=drawColor,line width= 0.7pt,line join=round] (554.37, 62.13) --
	(554.37,154.36);

\path[draw=drawColor,line width= 0.7pt,line join=round] (555.66, 62.13) --
	(555.66,154.36);

\path[draw=drawColor,line width= 0.7pt,line join=round] (556.96, 62.13) --
	(556.96,154.36);

\path[draw=drawColor,line width= 0.7pt,line join=round] (558.25, 62.13) --
	(558.25,154.36);

\path[draw=drawColor,line width= 0.7pt,line join=round] (559.54, 62.13) --
	(559.54,154.36);

\path[draw=drawColor,line width= 0.7pt,line join=round] (560.83, 62.13) --
	(560.83,154.36);

\path[draw=drawColor,line width= 0.7pt,line join=round] (562.12, 62.13) --
	(562.12,154.36);

\path[draw=drawColor,line width= 0.7pt,line join=round] (563.41, 62.13) --
	(563.41,154.36);
\definecolor{drawColor}{RGB}{211,211,211}

\path[draw=drawColor,line width= 0.7pt,line join=round] (429.12, 66.32) --
	(571.16, 66.32);

\path[draw=drawColor,line width= 0.7pt,line join=round] (429.12, 74.71) --
	(571.16, 74.71);

\path[draw=drawColor,line width= 0.7pt,line join=round] (429.12, 87.28) --
	(571.16, 87.28);

\path[draw=drawColor,line width= 0.7pt,line join=round] (429.12,108.25) --
	(571.16,108.25);

\path[draw=drawColor,line width= 0.7pt,line join=round] (429.12,129.21) --
	(571.16,129.21);

\path[draw=drawColor,line width= 0.7pt,line join=round] (429.12,141.79) --
	(571.16,141.79);

\path[draw=drawColor,line width= 0.7pt,line join=round] (429.12,150.17) --
	(571.16,150.17);

\path[draw=drawColor,line width= 0.7pt,line join=round] (435.57, 62.13) --
	(435.57,154.36);

\path[draw=drawColor,line width= 0.7pt,line join=round] (448.49, 62.13) --
	(448.49,154.36);

\path[draw=drawColor,line width= 0.7pt,line join=round] (467.86, 62.13) --
	(467.86,154.36);

\path[draw=drawColor,line width= 0.7pt,line join=round] (500.14, 62.13) --
	(500.14,154.36);

\path[draw=drawColor,line width= 0.7pt,line join=round] (532.42, 62.13) --
	(532.42,154.36);

\path[draw=drawColor,line width= 0.7pt,line join=round] (551.79, 62.13) --
	(551.79,154.36);

\path[draw=drawColor,line width= 0.7pt,line join=round] (564.70, 62.13) --
	(564.70,154.36);
\definecolor{drawColor}{RGB}{37,43,122}

\path[draw=drawColor,line width= 1.1pt,line join=round] (435.57, 66.32) --
	(435.75,106.81) --
	(435.93,117.14) --
	(435.98,117.61) --
	(436.02,119.10) --
	(436.06,120.16) --
	(436.14,123.32) --
	(436.23,125.56) --
	(436.35,126.70) --
	(436.44,127.58) --
	(437.29,130.83) --
	(438.22,138.68) --
	(439.18,142.24) --
	(439.95,144.09) --
	(440.58,145.42) --
	(441.29,146.04) --
	(442.27,146.42) --
	(443.28,147.23) --
	(444.64,147.58) --
	(446.75,147.97) --
	(453.09,148.62) --
	(564.70,150.17);
\definecolor{drawColor}{RGB}{211,211,211}

\path[draw=drawColor,line width= 0.6pt,line join=round] (429.12, 62.13) -- (571.16,154.36);
\end{scope}
\begin{scope}
\path[clip] (  0.00,  0.00) rectangle (578.16,180.67);
\definecolor{drawColor}{RGB}{0,0,0}

\path[draw=drawColor,line width= 0.6pt,line join=round,line cap=rect] (429.12, 62.13) --
	(429.12,154.36);
\end{scope}
\begin{scope}
\path[clip] (  0.00,  0.00) rectangle (578.16,180.67);
\definecolor{drawColor}{RGB}{0,0,0}

\node[text=drawColor,anchor=base east,inner sep=0pt, outer sep=0pt, scale=  0.50] at (422.62, 64.60) {0.00};

\node[text=drawColor,anchor=base east,inner sep=0pt, outer sep=0pt, scale=  0.50] at (422.62, 72.99) {0.10};

\node[text=drawColor,anchor=base east,inner sep=0pt, outer sep=0pt, scale=  0.50] at (422.62, 85.56) {0.25};

\node[text=drawColor,anchor=base east,inner sep=0pt, outer sep=0pt, scale=  0.50] at (422.62,106.52) {0.50};

\node[text=drawColor,anchor=base east,inner sep=0pt, outer sep=0pt, scale=  0.50] at (422.62,127.49) {0.75};

\node[text=drawColor,anchor=base east,inner sep=0pt, outer sep=0pt, scale=  0.50] at (422.62,140.06) {0.90};

\node[text=drawColor,anchor=base east,inner sep=0pt, outer sep=0pt, scale=  0.50] at (422.62,148.45) {1.00};
\end{scope}
\begin{scope}
\path[clip] (  0.00,  0.00) rectangle (578.16,180.67);
\definecolor{drawColor}{RGB}{0,0,0}

\path[draw=drawColor,line width= 0.6pt,line join=round] (425.62, 66.32) --
	(429.12, 66.32);

\path[draw=drawColor,line width= 0.6pt,line join=round] (425.62, 74.71) --
	(429.12, 74.71);

\path[draw=drawColor,line width= 0.6pt,line join=round] (425.62, 87.28) --
	(429.12, 87.28);

\path[draw=drawColor,line width= 0.6pt,line join=round] (425.62,108.25) --
	(429.12,108.25);

\path[draw=drawColor,line width= 0.6pt,line join=round] (425.62,129.21) --
	(429.12,129.21);

\path[draw=drawColor,line width= 0.6pt,line join=round] (425.62,141.79) --
	(429.12,141.79);

\path[draw=drawColor,line width= 0.6pt,line join=round] (425.62,150.17) --
	(429.12,150.17);
\end{scope}
\begin{scope}
\path[clip] (  0.00,  0.00) rectangle (578.16,180.67);
\definecolor{drawColor}{RGB}{0,0,0}

\path[draw=drawColor,line width= 0.6pt,line join=round,line cap=rect] (429.12, 62.13) --
	(571.16, 62.13);
\end{scope}
\begin{scope}
\path[clip] (  0.00,  0.00) rectangle (578.16,180.67);
\definecolor{drawColor}{RGB}{0,0,0}

\path[draw=drawColor,line width= 0.6pt,line join=round] (435.57, 58.63) --
	(435.57, 62.13);

\path[draw=drawColor,line width= 0.6pt,line join=round] (448.49, 58.63) --
	(448.49, 62.13);

\path[draw=drawColor,line width= 0.6pt,line join=round] (467.86, 58.63) --
	(467.86, 62.13);

\path[draw=drawColor,line width= 0.6pt,line join=round] (500.14, 58.63) --
	(500.14, 62.13);

\path[draw=drawColor,line width= 0.6pt,line join=round] (532.42, 58.63) --
	(532.42, 62.13);

\path[draw=drawColor,line width= 0.6pt,line join=round] (551.79, 58.63) --
	(551.79, 62.13);

\path[draw=drawColor,line width= 0.6pt,line join=round] (564.70, 58.63) --
	(564.70, 62.13);
\end{scope}
\begin{scope}
\path[clip] (  0.00,  0.00) rectangle (578.16,180.67);
\definecolor{drawColor}{RGB}{0,0,0}

\node[text=drawColor,anchor=base,inner sep=0pt, outer sep=0pt, scale=  0.50] at (435.57, 52.19) {0.00};

\node[text=drawColor,anchor=base,inner sep=0pt, outer sep=0pt, scale=  0.50] at (448.49, 52.19) {0.10};

\node[text=drawColor,anchor=base,inner sep=0pt, outer sep=0pt, scale=  0.50] at (467.86, 52.19) {0.25};

\node[text=drawColor,anchor=base,inner sep=0pt, outer sep=0pt, scale=  0.50] at (500.14, 52.19) {0.50};

\node[text=drawColor,anchor=base,inner sep=0pt, outer sep=0pt, scale=  0.50] at (532.42, 52.19) {0.75};

\node[text=drawColor,anchor=base,inner sep=0pt, outer sep=0pt, scale=  0.50] at (551.79, 52.19) {0.90};

\node[text=drawColor,anchor=base,inner sep=0pt, outer sep=0pt, scale=  0.50] at (564.70, 52.19) {1.00};
\end{scope}
\begin{scope}
\path[clip] (  0.00,  0.00) rectangle (578.16,180.67);
\definecolor{drawColor}{RGB}{0,0,0}

\node[text=drawColor,anchor=base,inner sep=0pt, outer sep=0pt, scale=  1.40] at (500.14, 27.75) {False positive fraction};
\end{scope}
\begin{scope}
\path[clip] (  0.00,  0.00) rectangle (578.16,180.67);
\definecolor{drawColor}{RGB}{0,0,0}

\node[text=drawColor,rotate= 90.00,anchor=base,inner sep=0pt, outer sep=0pt, scale=  1.40] at (406.41,108.25) {True positive fraction};
\end{scope}
\begin{scope}
\path[clip] (  0.00,  0.00) rectangle (578.16,180.67);
\definecolor{drawColor}{RGB}{0,0,0}

\node[text=drawColor,anchor=base,inner sep=0pt, outer sep=0pt, scale=  1.40] at (500.14,162.69) {\bfseries ROC Curve};
\end{scope}
\begin{scope}
\path[clip] (  0.00,  0.00) rectangle (578.16,180.67);
\definecolor{drawColor}{RGB}{0,0,0}

\node[text=drawColor,anchor=base west,inner sep=0pt, outer sep=0pt, scale=  1.40] at (193.71,  3.39) {Phase};
\end{scope}
\begin{scope}
\path[clip] (  0.00,  0.00) rectangle (578.16,180.67);
\definecolor{drawColor}{RGB}{134,182,178}

\path[draw=drawColor,line width= 1.7pt,line join=round] (242.00,  8.21) -- (253.56,  8.21);
\end{scope}
\begin{scope}
\path[clip] (  0.00,  0.00) rectangle (578.16,180.67);
\definecolor{drawColor}{RGB}{149,91,153}

\path[draw=drawColor,line width= 1.7pt,line join=round] (312.39,  8.21) -- (323.96,  8.21);
\end{scope}
\begin{scope}
\path[clip] (  0.00,  0.00) rectangle (578.16,180.67);
\definecolor{drawColor}{RGB}{0,0,0}

\node[text=drawColor,anchor=base west,inner sep=0pt, outer sep=0pt, scale=  1.20] at (262.01,  4.08) {Training};
\end{scope}
\begin{scope}
\path[clip] (  0.00,  0.00) rectangle (578.16,180.67);
\definecolor{drawColor}{RGB}{0,0,0}

\node[text=drawColor,anchor=base west,inner sep=0pt, outer sep=0pt, scale=  1.20] at (332.40,  4.08) {Validation};
\end{scope}
\end{tikzpicture}

%% file: figs/figure_2.tex
\begin{tikzpicture}[x=1pt,y=1pt]
\definecolor{fillColor}{RGB}{255,255,255}
\path[use as bounding box,fill=fillColor,fill opacity=0.00] (0,0) rectangle (578.16,180.67);
\begin{scope}
\path[clip] ( 50.12, 62.13) rectangle (185.72,154.36);
\definecolor{drawColor}{RGB}{134,182,178}

\path[draw=drawColor,line width= 1.7pt,line join=round] ( 56.28, 94.66) --
	( 62.45, 96.56) --
	( 68.61, 97.88) --
	( 74.77, 99.14) --
	( 80.94,100.47) --
	( 87.10,101.10) --
	( 93.27,101.69) --
	( 99.43,103.43) --
	(105.59,103.99) --
	(111.76,106.59) --
	(117.92,107.93) --
	(124.08,109.97) --
	(130.25,112.93) --
	(136.41,114.87) --
	(142.57,117.87) --
	(148.74,121.29) --
	(154.90,124.92) --
	(161.07,127.05) --
	(167.23,129.93) --
	(173.39,132.08) --
	(179.56,135.25);
\definecolor{drawColor}{RGB}{149,91,153}

\path[draw=drawColor,line width= 1.7pt,line join=round] ( 56.28, 96.12) --
	( 62.45, 97.80) --
	( 68.61, 99.29) --
	( 74.77, 99.81) --
	( 80.94, 99.95) --
	( 87.10,100.20) --
	( 93.27,100.13) --
	( 99.43,102.23) --
	(105.59,102.46) --
	(111.76,102.49) --
	(117.92,103.90) --
	(124.08,104.36) --
	(130.25,104.14) --
	(136.41,105.48) --
	(142.57,105.40) --
	(148.74,104.78) --
	(154.90,105.33) --
	(161.07,103.11) --
	(167.23,103.90) --
	(173.39,104.93) --
	(179.56,104.02);
\end{scope}
\begin{scope}
\path[clip] (  0.00,  0.00) rectangle (578.16,180.67);
\definecolor{drawColor}{RGB}{0,0,0}

\path[draw=drawColor,line width= 0.6pt,line join=round,line cap=rect] ( 50.12, 62.13) --
	( 50.12,154.36);
\end{scope}
\begin{scope}
\path[clip] (  0.00,  0.00) rectangle (578.16,180.67);
\definecolor{drawColor}{RGB}{0,0,0}

\node[text=drawColor,anchor=base east,inner sep=0pt, outer sep=0pt, scale=  1.20] at ( 43.62, 62.19) {0.5};

\node[text=drawColor,anchor=base east,inner sep=0pt, outer sep=0pt, scale=  1.20] at ( 43.62, 78.96) {0.6};

\node[text=drawColor,anchor=base east,inner sep=0pt, outer sep=0pt, scale=  1.20] at ( 43.62, 95.73) {0.7};

\node[text=drawColor,anchor=base east,inner sep=0pt, outer sep=0pt, scale=  1.20] at ( 43.62,112.50) {0.8};

\node[text=drawColor,anchor=base east,inner sep=0pt, outer sep=0pt, scale=  1.20] at ( 43.62,129.27) {0.9};

\node[text=drawColor,anchor=base east,inner sep=0pt, outer sep=0pt, scale=  1.20] at ( 43.62,146.04) {1.0};
\end{scope}
\begin{scope}
\path[clip] (  0.00,  0.00) rectangle (578.16,180.67);
\definecolor{drawColor}{RGB}{0,0,0}

\path[draw=drawColor,line width= 0.6pt,line join=round] ( 46.62, 66.32) --
	( 50.12, 66.32);

\path[draw=drawColor,line width= 0.6pt,line join=round] ( 46.62, 83.09) --
	( 50.12, 83.09);

\path[draw=drawColor,line width= 0.6pt,line join=round] ( 46.62, 99.86) --
	( 50.12, 99.86);

\path[draw=drawColor,line width= 0.6pt,line join=round] ( 46.62,116.63) --
	( 50.12,116.63);

\path[draw=drawColor,line width= 0.6pt,line join=round] ( 46.62,133.40) --
	( 50.12,133.40);

\path[draw=drawColor,line width= 0.6pt,line join=round] ( 46.62,150.17) --
	( 50.12,150.17);
\end{scope}
\begin{scope}
\path[clip] (  0.00,  0.00) rectangle (578.16,180.67);
\definecolor{drawColor}{RGB}{0,0,0}

\path[draw=drawColor,line width= 0.6pt,line join=round,line cap=rect] ( 50.12, 62.13) --
	(185.72, 62.13);
\end{scope}
\begin{scope}
\path[clip] (  0.00,  0.00) rectangle (578.16,180.67);
\definecolor{drawColor}{RGB}{0,0,0}

\path[draw=drawColor,line width= 0.6pt,line join=round] ( 50.12, 58.63) --
	( 50.12, 62.13);

\path[draw=drawColor,line width= 0.6pt,line join=round] ( 80.94, 58.63) --
	( 80.94, 62.13);

\path[draw=drawColor,line width= 0.6pt,line join=round] (111.76, 58.63) --
	(111.76, 62.13);

\path[draw=drawColor,line width= 0.6pt,line join=round] (142.57, 58.63) --
	(142.57, 62.13);

\path[draw=drawColor,line width= 0.6pt,line join=round] (173.39, 58.63) --
	(173.39, 62.13);
\end{scope}
\begin{scope}
\path[clip] (  0.00,  0.00) rectangle (578.16,180.67);
\definecolor{drawColor}{RGB}{0,0,0}

\node[text=drawColor,anchor=base,inner sep=0pt, outer sep=0pt, scale=  1.20] at ( 50.12, 47.37) {0};

\node[text=drawColor,anchor=base,inner sep=0pt, outer sep=0pt, scale=  1.20] at ( 80.94, 47.37) {5};

\node[text=drawColor,anchor=base,inner sep=0pt, outer sep=0pt, scale=  1.20] at (111.76, 47.37) {10};

\node[text=drawColor,anchor=base,inner sep=0pt, outer sep=0pt, scale=  1.20] at (142.57, 47.37) {15};

\node[text=drawColor,anchor=base,inner sep=0pt, outer sep=0pt, scale=  1.20] at (173.39, 47.37) {20};
\end{scope}
\begin{scope}
\path[clip] (  0.00,  0.00) rectangle (578.16,180.67);
\definecolor{drawColor}{RGB}{0,0,0}

\node[text=drawColor,anchor=base,inner sep=0pt, outer sep=0pt, scale=  1.40] at (117.92, 27.75) {Epoch};
\end{scope}
\begin{scope}
\path[clip] (  0.00,  0.00) rectangle (578.16,180.67);
\definecolor{drawColor}{RGB}{0,0,0}

\node[text=drawColor,rotate= 90.00,anchor=base,inner sep=0pt, outer sep=0pt, scale=  1.40] at ( 20.97,108.25) {Binary Accuracy};
\end{scope}
\begin{scope}
\path[clip] (  0.00,  0.00) rectangle (578.16,180.67);
\definecolor{drawColor}{RGB}{0,0,0}

\node[text=drawColor,anchor=base,inner sep=0pt, outer sep=0pt, scale=  1.40] at (117.92,162.69) {\bfseries Accuracy};
\end{scope}
\begin{scope}
\path[clip] (242.84, 62.13) rectangle (378.44,154.36);
\definecolor{drawColor}{RGB}{134,182,178}

\path[draw=drawColor,line width= 1.7pt,line join=round] (249.00,111.21) --
	(255.17,109.19) --
	(261.33,107.88) --
	(267.49,107.10) --
	(273.66,106.57) --
	(279.82,105.84) --
	(285.99,105.44) --
	(292.15,104.80) --
	(298.31,104.21) --
	(304.48,103.11) --
	(310.64,102.04) --
	(316.80,100.89) --
	(322.97, 99.29) --
	(329.13, 97.22) --
	(335.29, 95.41) --
	(341.46, 92.73) --
	(347.62, 89.98) --
	(353.79, 88.07) --
	(359.95, 85.75) --
	(366.11, 83.58) --
	(372.28, 81.46);
\definecolor{drawColor}{RGB}{149,91,153}

\path[draw=drawColor,line width= 1.7pt,line join=round] (249.00,110.76) --
	(255.17,107.89) --
	(261.33,106.89) --
	(267.49,107.22) --
	(273.66,106.56) --
	(279.82,106.45) --
	(285.99,106.57) --
	(292.15,106.04) --
	(298.31,105.57) --
	(304.48,105.24) --
	(310.64,105.06) --
	(316.80,105.13) --
	(322.97,105.15) --
	(329.13,104.38) --
	(335.29,106.34) --
	(341.46,106.67) --
	(347.62,111.18) --
	(353.79,113.14) --
	(359.95,111.06) --
	(366.11,114.74) --
	(372.28,114.33);
\end{scope}
\begin{scope}
\path[clip] (  0.00,  0.00) rectangle (578.16,180.67);
\definecolor{drawColor}{RGB}{0,0,0}

\path[draw=drawColor,line width= 0.6pt,line join=round,line cap=rect] (242.84, 62.13) --
	(242.84,154.36);
\end{scope}
\begin{scope}
\path[clip] (  0.00,  0.00) rectangle (578.16,180.67);
\definecolor{drawColor}{RGB}{0,0,0}

\node[text=drawColor,anchor=base east,inner sep=0pt, outer sep=0pt, scale=  1.20] at (236.34, 62.19) {0.0};

\node[text=drawColor,anchor=base east,inner sep=0pt, outer sep=0pt, scale=  1.20] at (236.34, 83.15) {0.5};

\node[text=drawColor,anchor=base east,inner sep=0pt, outer sep=0pt, scale=  1.20] at (236.34,104.11) {1.0};

\node[text=drawColor,anchor=base east,inner sep=0pt, outer sep=0pt, scale=  1.20] at (236.34,125.08) {1.5};

\node[text=drawColor,anchor=base east,inner sep=0pt, outer sep=0pt, scale=  1.20] at (236.34,146.04) {2.0};
\end{scope}
\begin{scope}
\path[clip] (  0.00,  0.00) rectangle (578.16,180.67);
\definecolor{drawColor}{RGB}{0,0,0}

\path[draw=drawColor,line width= 0.6pt,line join=round] (239.34, 66.32) --
	(242.84, 66.32);

\path[draw=drawColor,line width= 0.6pt,line join=round] (239.34, 87.28) --
	(242.84, 87.28);

\path[draw=drawColor,line width= 0.6pt,line join=round] (239.34,108.25) --
	(242.84,108.25);

\path[draw=drawColor,line width= 0.6pt,line join=round] (239.34,129.21) --
	(242.84,129.21);

\path[draw=drawColor,line width= 0.6pt,line join=round] (239.34,150.17) --
	(242.84,150.17);
\end{scope}
\begin{scope}
\path[clip] (  0.00,  0.00) rectangle (578.16,180.67);
\definecolor{drawColor}{RGB}{0,0,0}

\path[draw=drawColor,line width= 0.6pt,line join=round,line cap=rect] (242.84, 62.13) --
	(378.44, 62.13);
\end{scope}
\begin{scope}
\path[clip] (  0.00,  0.00) rectangle (578.16,180.67);
\definecolor{drawColor}{RGB}{0,0,0}

\path[draw=drawColor,line width= 0.6pt,line join=round] (242.84, 58.63) --
	(242.84, 62.13);

\path[draw=drawColor,line width= 0.6pt,line join=round] (273.66, 58.63) --
	(273.66, 62.13);

\path[draw=drawColor,line width= 0.6pt,line join=round] (304.48, 58.63) --
	(304.48, 62.13);

\path[draw=drawColor,line width= 0.6pt,line join=round] (335.29, 58.63) --
	(335.29, 62.13);

\path[draw=drawColor,line width= 0.6pt,line join=round] (366.11, 58.63) --
	(366.11, 62.13);
\end{scope}
\begin{scope}
\path[clip] (  0.00,  0.00) rectangle (578.16,180.67);
\definecolor{drawColor}{RGB}{0,0,0}

\node[text=drawColor,anchor=base,inner sep=0pt, outer sep=0pt, scale=  1.20] at (242.84, 47.37) {0};

\node[text=drawColor,anchor=base,inner sep=0pt, outer sep=0pt, scale=  1.20] at (273.66, 47.37) {5};

\node[text=drawColor,anchor=base,inner sep=0pt, outer sep=0pt, scale=  1.20] at (304.48, 47.37) {10};

\node[text=drawColor,anchor=base,inner sep=0pt, outer sep=0pt, scale=  1.20] at (335.29, 47.37) {15};

\node[text=drawColor,anchor=base,inner sep=0pt, outer sep=0pt, scale=  1.20] at (366.11, 47.37) {20};
\end{scope}
\begin{scope}
\path[clip] (  0.00,  0.00) rectangle (578.16,180.67);
\definecolor{drawColor}{RGB}{0,0,0}

\node[text=drawColor,anchor=base,inner sep=0pt, outer sep=0pt, scale=  1.40] at (310.64, 27.75) {Epoch};
\end{scope}
\begin{scope}
\path[clip] (  0.00,  0.00) rectangle (578.16,180.67);
\definecolor{drawColor}{RGB}{0,0,0}

\node[text=drawColor,rotate= 90.00,anchor=base,inner sep=0pt, outer sep=0pt, scale=  1.40] at (213.69,108.25) {Loss};
\end{scope}
\begin{scope}
\path[clip] (  0.00,  0.00) rectangle (578.16,180.67);
\definecolor{drawColor}{RGB}{0,0,0}

\node[text=drawColor,anchor=base,inner sep=0pt, outer sep=0pt, scale=  1.40] at (310.64,162.69) {\bfseries Loss};
\end{scope}
\begin{scope}
\path[clip] (429.12, 62.13) rectangle (571.16,154.36);
\definecolor{drawColor}{RGB}{255,255,255}

\path[draw=drawColor,line width= 0.7pt,line join=round] (429.12, 67.16) --
	(571.16, 67.16);

\path[draw=drawColor,line width= 0.7pt,line join=round] (429.12, 68.00) --
	(571.16, 68.00);

\path[draw=drawColor,line width= 0.7pt,line join=round] (429.12, 68.84) --
	(571.16, 68.84);

\path[draw=drawColor,line width= 0.7pt,line join=round] (429.12, 69.68) --
	(571.16, 69.68);

\path[draw=drawColor,line width= 0.7pt,line join=round] (429.12, 70.51) --
	(571.16, 70.51);

\path[draw=drawColor,line width= 0.7pt,line join=round] (429.12, 71.35) --
	(571.16, 71.35);

\path[draw=drawColor,line width= 0.7pt,line join=round] (429.12, 72.19) --
	(571.16, 72.19);

\path[draw=drawColor,line width= 0.7pt,line join=round] (429.12, 73.03) --
	(571.16, 73.03);

\path[draw=drawColor,line width= 0.7pt,line join=round] (429.12, 73.87) --
	(571.16, 73.87);

\path[draw=drawColor,line width= 0.7pt,line join=round] (429.12,142.62) --
	(571.16,142.62);

\path[draw=drawColor,line width= 0.7pt,line join=round] (429.12,143.46) --
	(571.16,143.46);

\path[draw=drawColor,line width= 0.7pt,line join=round] (429.12,144.30) --
	(571.16,144.30);

\path[draw=drawColor,line width= 0.7pt,line join=round] (429.12,145.14) --
	(571.16,145.14);

\path[draw=drawColor,line width= 0.7pt,line join=round] (429.12,145.98) --
	(571.16,145.98);

\path[draw=drawColor,line width= 0.7pt,line join=round] (429.12,146.82) --
	(571.16,146.82);

\path[draw=drawColor,line width= 0.7pt,line join=round] (429.12,147.66) --
	(571.16,147.66);

\path[draw=drawColor,line width= 0.7pt,line join=round] (429.12,148.49) --
	(571.16,148.49);

\path[draw=drawColor,line width= 0.7pt,line join=round] (429.12,149.33) --
	(571.16,149.33);

\path[draw=drawColor,line width= 0.7pt,line join=round] (436.86, 62.13) --
	(436.86,154.36);

\path[draw=drawColor,line width= 0.7pt,line join=round] (438.16, 62.13) --
	(438.16,154.36);

\path[draw=drawColor,line width= 0.7pt,line join=round] (439.45, 62.13) --
	(439.45,154.36);

\path[draw=drawColor,line width= 0.7pt,line join=round] (440.74, 62.13) --
	(440.74,154.36);

\path[draw=drawColor,line width= 0.7pt,line join=round] (442.03, 62.13) --
	(442.03,154.36);

\path[draw=drawColor,line width= 0.7pt,line join=round] (443.32, 62.13) --
	(443.32,154.36);

\path[draw=drawColor,line width= 0.7pt,line join=round] (444.61, 62.13) --
	(444.61,154.36);

\path[draw=drawColor,line width= 0.7pt,line join=round] (445.90, 62.13) --
	(445.90,154.36);

\path[draw=drawColor,line width= 0.7pt,line join=round] (447.19, 62.13) --
	(447.19,154.36);

\path[draw=drawColor,line width= 0.7pt,line join=round] (553.08, 62.13) --
	(553.08,154.36);

\path[draw=drawColor,line width= 0.7pt,line join=round] (554.37, 62.13) --
	(554.37,154.36);

\path[draw=drawColor,line width= 0.7pt,line join=round] (555.66, 62.13) --
	(555.66,154.36);

\path[draw=drawColor,line width= 0.7pt,line join=round] (556.96, 62.13) --
	(556.96,154.36);

\path[draw=drawColor,line width= 0.7pt,line join=round] (558.25, 62.13) --
	(558.25,154.36);

\path[draw=drawColor,line width= 0.7pt,line join=round] (559.54, 62.13) --
	(559.54,154.36);

\path[draw=drawColor,line width= 0.7pt,line join=round] (560.83, 62.13) --
	(560.83,154.36);

\path[draw=drawColor,line width= 0.7pt,line join=round] (562.12, 62.13) --
	(562.12,154.36);

\path[draw=drawColor,line width= 0.7pt,line join=round] (563.41, 62.13) --
	(563.41,154.36);
\definecolor{drawColor}{RGB}{211,211,211}

\path[draw=drawColor,line width= 0.7pt,line join=round] (429.12, 66.32) --
	(571.16, 66.32);

\path[draw=drawColor,line width= 0.7pt,line join=round] (429.12, 74.71) --
	(571.16, 74.71);

\path[draw=drawColor,line width= 0.7pt,line join=round] (429.12, 87.28) --
	(571.16, 87.28);

\path[draw=drawColor,line width= 0.7pt,line join=round] (429.12,108.25) --
	(571.16,108.25);

\path[draw=drawColor,line width= 0.7pt,line join=round] (429.12,129.21) --
	(571.16,129.21);

\path[draw=drawColor,line width= 0.7pt,line join=round] (429.12,141.79) --
	(571.16,141.79);

\path[draw=drawColor,line width= 0.7pt,line join=round] (429.12,150.17) --
	(571.16,150.17);

\path[draw=drawColor,line width= 0.7pt,line join=round] (435.57, 62.13) --
	(435.57,154.36);

\path[draw=drawColor,line width= 0.7pt,line join=round] (448.49, 62.13) --
	(448.49,154.36);

\path[draw=drawColor,line width= 0.7pt,line join=round] (467.86, 62.13) --
	(467.86,154.36);

\path[draw=drawColor,line width= 0.7pt,line join=round] (500.14, 62.13) --
	(500.14,154.36);

\path[draw=drawColor,line width= 0.7pt,line join=round] (532.42, 62.13) --
	(532.42,154.36);

\path[draw=drawColor,line width= 0.7pt,line join=round] (551.79, 62.13) --
	(551.79,154.36);

\path[draw=drawColor,line width= 0.7pt,line join=round] (564.70, 62.13) --
	(564.70,154.36);
\definecolor{drawColor}{RGB}{37,43,122}

\path[draw=drawColor,line width= 1.1pt,line join=round] (435.57, 66.32) --
	(436.26, 73.17) --
	(440.23, 83.90) --
	(443.15, 88.82) --
	(445.20, 93.25) --
	(447.93, 96.48) --
	(450.53,100.18) --
	(453.02,103.01) --
	(455.07,105.42) --
	(458.05,108.17) --
	(460.47,110.58) --
	(463.08,112.60) --
	(465.81,114.62) --
	(468.54,117.20) --
	(471.70,119.21) --
	(474.50,120.99) --
	(478.78,123.24) --
	(484.43,127.35) --
	(492.56,131.63) --
	(506.28,137.35) --
	(536.95,147.51) --
	(564.70,150.17);
\definecolor{drawColor}{RGB}{211,211,211}

\path[draw=drawColor,line width= 0.6pt,line join=round] (429.12, 62.13) -- (571.16,154.36);
\end{scope}
\begin{scope}
\path[clip] (  0.00,  0.00) rectangle (578.16,180.67);
\definecolor{drawColor}{RGB}{0,0,0}

\path[draw=drawColor,line width= 0.6pt,line join=round,line cap=rect] (429.12, 62.13) --
	(429.12,154.36);
\end{scope}
\begin{scope}
\path[clip] (  0.00,  0.00) rectangle (578.16,180.67);
\definecolor{drawColor}{RGB}{0,0,0}

\node[text=drawColor,anchor=base east,inner sep=0pt, outer sep=0pt, scale=  0.50] at (422.62, 64.60) {0.00};

\node[text=drawColor,anchor=base east,inner sep=0pt, outer sep=0pt, scale=  0.50] at (422.62, 72.99) {0.10};

\node[text=drawColor,anchor=base east,inner sep=0pt, outer sep=0pt, scale=  0.50] at (422.62, 85.56) {0.25};

\node[text=drawColor,anchor=base east,inner sep=0pt, outer sep=0pt, scale=  0.50] at (422.62,106.52) {0.50};

\node[text=drawColor,anchor=base east,inner sep=0pt, outer sep=0pt, scale=  0.50] at (422.62,127.49) {0.75};

\node[text=drawColor,anchor=base east,inner sep=0pt, outer sep=0pt, scale=  0.50] at (422.62,140.06) {0.90};

\node[text=drawColor,anchor=base east,inner sep=0pt, outer sep=0pt, scale=  0.50] at (422.62,148.45) {1.00};
\end{scope}
\begin{scope}
\path[clip] (  0.00,  0.00) rectangle (578.16,180.67);
\definecolor{drawColor}{RGB}{0,0,0}

\path[draw=drawColor,line width= 0.6pt,line join=round] (425.62, 66.32) --
	(429.12, 66.32);

\path[draw=drawColor,line width= 0.6pt,line join=round] (425.62, 74.71) --
	(429.12, 74.71);

\path[draw=drawColor,line width= 0.6pt,line join=round] (425.62, 87.28) --
	(429.12, 87.28);

\path[draw=drawColor,line width= 0.6pt,line join=round] (425.62,108.25) --
	(429.12,108.25);

\path[draw=drawColor,line width= 0.6pt,line join=round] (425.62,129.21) --
	(429.12,129.21);

\path[draw=drawColor,line width= 0.6pt,line join=round] (425.62,141.79) --
	(429.12,141.79);

\path[draw=drawColor,line width= 0.6pt,line join=round] (425.62,150.17) --
	(429.12,150.17);
\end{scope}
\begin{scope}
\path[clip] (  0.00,  0.00) rectangle (578.16,180.67);
\definecolor{drawColor}{RGB}{0,0,0}

\path[draw=drawColor,line width= 0.6pt,line join=round,line cap=rect] (429.12, 62.13) --
	(571.16, 62.13);
\end{scope}
\begin{scope}
\path[clip] (  0.00,  0.00) rectangle (578.16,180.67);
\definecolor{drawColor}{RGB}{0,0,0}

\path[draw=drawColor,line width= 0.6pt,line join=round] (435.57, 58.63) --
	(435.57, 62.13);

\path[draw=drawColor,line width= 0.6pt,line join=round] (448.49, 58.63) --
	(448.49, 62.13);

\path[draw=drawColor,line width= 0.6pt,line join=round] (467.86, 58.63) --
	(467.86, 62.13);

\path[draw=drawColor,line width= 0.6pt,line join=round] (500.14, 58.63) --
	(500.14, 62.13);

\path[draw=drawColor,line width= 0.6pt,line join=round] (532.42, 58.63) --
	(532.42, 62.13);

\path[draw=drawColor,line width= 0.6pt,line join=round] (551.79, 58.63) --
	(551.79, 62.13);

\path[draw=drawColor,line width= 0.6pt,line join=round] (564.70, 58.63) --
	(564.70, 62.13);
\end{scope}
\begin{scope}
\path[clip] (  0.00,  0.00) rectangle (578.16,180.67);
\definecolor{drawColor}{RGB}{0,0,0}

\node[text=drawColor,anchor=base,inner sep=0pt, outer sep=0pt, scale=  0.50] at (435.57, 52.19) {0.00};

\node[text=drawColor,anchor=base,inner sep=0pt, outer sep=0pt, scale=  0.50] at (448.49, 52.19) {0.10};

\node[text=drawColor,anchor=base,inner sep=0pt, outer sep=0pt, scale=  0.50] at (467.86, 52.19) {0.25};

\node[text=drawColor,anchor=base,inner sep=0pt, outer sep=0pt, scale=  0.50] at (500.14, 52.19) {0.50};

\node[text=drawColor,anchor=base,inner sep=0pt, outer sep=0pt, scale=  0.50] at (532.42, 52.19) {0.75};

\node[text=drawColor,anchor=base,inner sep=0pt, outer sep=0pt, scale=  0.50] at (551.79, 52.19) {0.90};

\node[text=drawColor,anchor=base,inner sep=0pt, outer sep=0pt, scale=  0.50] at (564.70, 52.19) {1.00};
\end{scope}
\begin{scope}
\path[clip] (  0.00,  0.00) rectangle (578.16,180.67);
\definecolor{drawColor}{RGB}{0,0,0}

\node[text=drawColor,anchor=base,inner sep=0pt, outer sep=0pt, scale=  1.40] at (500.14, 27.75) {False positive fraction};
\end{scope}
\begin{scope}
\path[clip] (  0.00,  0.00) rectangle (578.16,180.67);
\definecolor{drawColor}{RGB}{0,0,0}

\node[text=drawColor,rotate= 90.00,anchor=base,inner sep=0pt, outer sep=0pt, scale=  1.40] at (406.41,108.25) {True positive fraction};
\end{scope}
\begin{scope}
\path[clip] (  0.00,  0.00) rectangle (578.16,180.67);
\definecolor{drawColor}{RGB}{0,0,0}

\node[text=drawColor,anchor=base,inner sep=0pt, outer sep=0pt, scale=  1.40] at (500.14,162.69) {\bfseries ROC Curve};
\end{scope}
\begin{scope}
\path[clip] (  0.00,  0.00) rectangle (578.16,180.67);
\definecolor{drawColor}{RGB}{0,0,0}

\node[text=drawColor,anchor=base west,inner sep=0pt, outer sep=0pt, scale=  1.40] at (193.71,  3.39) {Phase};
\end{scope}
\begin{scope}
\path[clip] (  0.00,  0.00) rectangle (578.16,180.67);
\definecolor{drawColor}{RGB}{134,182,178}

\path[draw=drawColor,line width= 1.7pt,line join=round] (242.00,  8.21) -- (253.56,  8.21);
\end{scope}
\begin{scope}
\path[clip] (  0.00,  0.00) rectangle (578.16,180.67);
\definecolor{drawColor}{RGB}{149,91,153}

\path[draw=drawColor,line width= 1.7pt,line join=round] (312.39,  8.21) -- (323.96,  8.21);
\end{scope}
\begin{scope}
\path[clip] (  0.00,  0.00) rectangle (578.16,180.67);
\definecolor{drawColor}{RGB}{0,0,0}

\node[text=drawColor,anchor=base west,inner sep=0pt, outer sep=0pt, scale=  1.20] at (262.01,  4.08) {Training};
\end{scope}
\begin{scope}
\path[clip] (  0.00,  0.00) rectangle (578.16,180.67);
\definecolor{drawColor}{RGB}{0,0,0}

\node[text=drawColor,anchor=base west,inner sep=0pt, outer sep=0pt, scale=  1.20] at (332.40,  4.08) {Validation};
\end{scope}
\end{tikzpicture}

%% file: figs/figure_3.tex
\begin{tikzpicture}[x=1pt,y=1pt]
\definecolor{fillColor}{RGB}{255,255,255}
\path[use as bounding box,fill=fillColor,fill opacity=0.00] (0,0) rectangle (578.16,180.67);
\begin{scope}
\path[clip] ( 50.12, 62.13) rectangle (185.72,154.36);
\definecolor{drawColor}{RGB}{134,182,178}

\path[draw=drawColor,line width= 1.7pt,line join=round] ( 56.28, 96.41) --
	( 62.77,111.38) --
	( 69.26,123.51) --
	( 75.75,126.49) --
	( 82.24,130.67) --
	( 88.72,134.37) --
	( 95.21,133.74) --
	(101.70,136.84) --
	(108.19,137.56) --
	(114.68,139.91) --
	(121.16,140.39) --
	(127.65,143.25) --
	(134.14,142.07) --
	(140.63,143.75) --
	(147.12,144.69) --
	(153.60,146.26) --
	(160.09,146.62) --
	(166.58,147.32) --
	(173.07,145.93) --
	(179.56,148.02);
\definecolor{drawColor}{RGB}{149,91,153}

\path[draw=drawColor,line width= 1.7pt,line join=round] ( 56.28,101.09) --
	( 62.77,110.21) --
	( 69.26,118.81) --
	( 75.75,127.53) --
	( 82.24,126.64) --
	( 88.72,135.70) --
	( 95.21,137.31) --
	(101.70,130.89) --
	(108.19,138.10) --
	(114.68,133.90) --
	(121.16,140.66) --
	(127.65,140.39) --
	(134.14,143.51) --
	(140.63,143.97) --
	(147.12,143.85) --
	(153.60,145.53) --
	(160.09,145.76) --
	(166.58,145.64) --
	(173.07,145.31) --
	(179.56,146.15);
\end{scope}
\begin{scope}
\path[clip] (  0.00,  0.00) rectangle (578.16,180.67);
\definecolor{drawColor}{RGB}{0,0,0}

\path[draw=drawColor,line width= 0.6pt,line join=round,line cap=rect] ( 50.12, 62.13) --
	( 50.12,154.36);
\end{scope}
\begin{scope}
\path[clip] (  0.00,  0.00) rectangle (578.16,180.67);
\definecolor{drawColor}{RGB}{0,0,0}

\node[text=drawColor,anchor=base east,inner sep=0pt, outer sep=0pt, scale=  1.20] at ( 43.62, 62.19) {0.5};

\node[text=drawColor,anchor=base east,inner sep=0pt, outer sep=0pt, scale=  1.20] at ( 43.62, 78.96) {0.6};

\node[text=drawColor,anchor=base east,inner sep=0pt, outer sep=0pt, scale=  1.20] at ( 43.62, 95.73) {0.7};

\node[text=drawColor,anchor=base east,inner sep=0pt, outer sep=0pt, scale=  1.20] at ( 43.62,112.50) {0.8};

\node[text=drawColor,anchor=base east,inner sep=0pt, outer sep=0pt, scale=  1.20] at ( 43.62,129.27) {0.9};

\node[text=drawColor,anchor=base east,inner sep=0pt, outer sep=0pt, scale=  1.20] at ( 43.62,146.04) {1.0};
\end{scope}
\begin{scope}
\path[clip] (  0.00,  0.00) rectangle (578.16,180.67);
\definecolor{drawColor}{RGB}{0,0,0}

\path[draw=drawColor,line width= 0.6pt,line join=round] ( 46.62, 66.32) --
	( 50.12, 66.32);

\path[draw=drawColor,line width= 0.6pt,line join=round] ( 46.62, 83.09) --
	( 50.12, 83.09);

\path[draw=drawColor,line width= 0.6pt,line join=round] ( 46.62, 99.86) --
	( 50.12, 99.86);

\path[draw=drawColor,line width= 0.6pt,line join=round] ( 46.62,116.63) --
	( 50.12,116.63);

\path[draw=drawColor,line width= 0.6pt,line join=round] ( 46.62,133.40) --
	( 50.12,133.40);

\path[draw=drawColor,line width= 0.6pt,line join=round] ( 46.62,150.17) --
	( 50.12,150.17);
\end{scope}
\begin{scope}
\path[clip] (  0.00,  0.00) rectangle (578.16,180.67);
\definecolor{drawColor}{RGB}{0,0,0}

\path[draw=drawColor,line width= 0.6pt,line join=round,line cap=rect] ( 50.12, 62.13) --
	(185.72, 62.13);
\end{scope}
\begin{scope}
\path[clip] (  0.00,  0.00) rectangle (578.16,180.67);
\definecolor{drawColor}{RGB}{0,0,0}

\path[draw=drawColor,line width= 0.6pt,line join=round] ( 82.24, 58.63) --
	( 82.24, 62.13);

\path[draw=drawColor,line width= 0.6pt,line join=round] (114.68, 58.63) --
	(114.68, 62.13);

\path[draw=drawColor,line width= 0.6pt,line join=round] (147.12, 58.63) --
	(147.12, 62.13);

\path[draw=drawColor,line width= 0.6pt,line join=round] (179.56, 58.63) --
	(179.56, 62.13);
\end{scope}
\begin{scope}
\path[clip] (  0.00,  0.00) rectangle (578.16,180.67);
\definecolor{drawColor}{RGB}{0,0,0}

\node[text=drawColor,anchor=base,inner sep=0pt, outer sep=0pt, scale=  1.20] at ( 82.24, 47.37) {5};

\node[text=drawColor,anchor=base,inner sep=0pt, outer sep=0pt, scale=  1.20] at (114.68, 47.37) {10};

\node[text=drawColor,anchor=base,inner sep=0pt, outer sep=0pt, scale=  1.20] at (147.12, 47.37) {15};

\node[text=drawColor,anchor=base,inner sep=0pt, outer sep=0pt, scale=  1.20] at (179.56, 47.37) {20};
\end{scope}
\begin{scope}
\path[clip] (  0.00,  0.00) rectangle (578.16,180.67);
\definecolor{drawColor}{RGB}{0,0,0}

\node[text=drawColor,anchor=base,inner sep=0pt, outer sep=0pt, scale=  1.40] at (117.92, 27.75) {Epoch};
\end{scope}
\begin{scope}
\path[clip] (  0.00,  0.00) rectangle (578.16,180.67);
\definecolor{drawColor}{RGB}{0,0,0}

\node[text=drawColor,rotate= 90.00,anchor=base,inner sep=0pt, outer sep=0pt, scale=  1.40] at ( 20.97,108.25) {Binary Accuracy};
\end{scope}
\begin{scope}
\path[clip] (  0.00,  0.00) rectangle (578.16,180.67);
\definecolor{drawColor}{RGB}{0,0,0}

\node[text=drawColor,anchor=base,inner sep=0pt, outer sep=0pt, scale=  1.40] at (117.92,162.69) {\bfseries Accuracy};
\end{scope}
\begin{scope}
\path[clip] (233.51, 62.13) rectangle (378.44,154.36);
\definecolor{drawColor}{RGB}{134,182,178}

\path[draw=drawColor,line width= 1.7pt,line join=round] (240.10, 76.22) --
	(247.03, 74.17) --
	(253.97, 72.37) --
	(260.90, 71.78) --
	(267.83, 71.18) --
	(274.77, 70.38) --
	(281.70, 70.50) --
	(288.64, 69.86) --
	(295.57, 69.72) --
	(302.51, 69.01) --
	(309.44, 68.81) --
	(316.38, 68.33) --
	(323.31, 68.46) --
	(330.25, 68.09) --
	(337.18, 67.89) --
	(344.11, 67.40) --
	(351.05, 67.33) --
	(357.98, 67.20) --
	(364.92, 67.50) --
	(371.85, 66.98);
\definecolor{drawColor}{RGB}{149,91,153}

\path[draw=drawColor,line width= 1.7pt,line join=round] (240.10, 74.96) --
	(247.03, 74.26) --
	(253.97, 72.95) --
	(260.90, 71.49) --
	(267.83, 71.31) --
	(274.77, 69.82) --
	(281.70, 69.71) --
	(288.64, 70.98) --
	(295.57, 69.56) --
	(302.51, 70.07) --
	(309.44, 68.79) --
	(316.38, 69.20) --
	(323.31, 68.21) --
	(330.25, 68.21) --
	(337.18, 68.10) --
	(344.11, 67.70) --
	(351.05, 67.67) --
	(357.98, 67.66) --
	(364.92, 68.02) --
	(371.85, 68.01);
\end{scope}
\begin{scope}
\path[clip] (  0.00,  0.00) rectangle (578.16,180.67);
\definecolor{drawColor}{RGB}{0,0,0}

\path[draw=drawColor,line width= 0.6pt,line join=round,line cap=rect] (233.51, 62.13) --
	(233.51,154.36);
\end{scope}
\begin{scope}
\path[clip] (  0.00,  0.00) rectangle (578.16,180.67);
\definecolor{drawColor}{RGB}{0,0,0}

\node[text=drawColor,anchor=base east,inner sep=0pt, outer sep=0pt, scale=  1.20] at (227.01, 62.19) {0};

\node[text=drawColor,anchor=base east,inner sep=0pt, outer sep=0pt, scale=  1.20] at (227.01, 78.96) {1};

\node[text=drawColor,anchor=base east,inner sep=0pt, outer sep=0pt, scale=  1.20] at (227.01, 95.73) {2};

\node[text=drawColor,anchor=base east,inner sep=0pt, outer sep=0pt, scale=  1.20] at (227.01,112.50) {3};

\node[text=drawColor,anchor=base east,inner sep=0pt, outer sep=0pt, scale=  1.20] at (227.01,129.27) {4};

\node[text=drawColor,anchor=base east,inner sep=0pt, outer sep=0pt, scale=  1.20] at (227.01,146.04) {5};
\end{scope}
\begin{scope}
\path[clip] (  0.00,  0.00) rectangle (578.16,180.67);
\definecolor{drawColor}{RGB}{0,0,0}

\path[draw=drawColor,line width= 0.6pt,line join=round] (230.01, 66.32) --
	(233.51, 66.32);

\path[draw=drawColor,line width= 0.6pt,line join=round] (230.01, 83.09) --
	(233.51, 83.09);

\path[draw=drawColor,line width= 0.6pt,line join=round] (230.01, 99.86) --
	(233.51, 99.86);

\path[draw=drawColor,line width= 0.6pt,line join=round] (230.01,116.63) --
	(233.51,116.63);

\path[draw=drawColor,line width= 0.6pt,line join=round] (230.01,133.40) --
	(233.51,133.40);

\path[draw=drawColor,line width= 0.6pt,line join=round] (230.01,150.17) --
	(233.51,150.17);
\end{scope}
\begin{scope}
\path[clip] (  0.00,  0.00) rectangle (578.16,180.67);
\definecolor{drawColor}{RGB}{0,0,0}

\path[draw=drawColor,line width= 0.6pt,line join=round,line cap=rect] (233.51, 62.13) --
	(378.44, 62.13);
\end{scope}
\begin{scope}
\path[clip] (  0.00,  0.00) rectangle (578.16,180.67);
\definecolor{drawColor}{RGB}{0,0,0}

\path[draw=drawColor,line width= 0.6pt,line join=round] (267.83, 58.63) --
	(267.83, 62.13);

\path[draw=drawColor,line width= 0.6pt,line join=round] (302.51, 58.63) --
	(302.51, 62.13);

\path[draw=drawColor,line width= 0.6pt,line join=round] (337.18, 58.63) --
	(337.18, 62.13);

\path[draw=drawColor,line width= 0.6pt,line join=round] (371.85, 58.63) --
	(371.85, 62.13);
\end{scope}
\begin{scope}
\path[clip] (  0.00,  0.00) rectangle (578.16,180.67);
\definecolor{drawColor}{RGB}{0,0,0}

\node[text=drawColor,anchor=base,inner sep=0pt, outer sep=0pt, scale=  1.20] at (267.83, 47.37) {5};

\node[text=drawColor,anchor=base,inner sep=0pt, outer sep=0pt, scale=  1.20] at (302.51, 47.37) {10};

\node[text=drawColor,anchor=base,inner sep=0pt, outer sep=0pt, scale=  1.20] at (337.18, 47.37) {15};

\node[text=drawColor,anchor=base,inner sep=0pt, outer sep=0pt, scale=  1.20] at (371.85, 47.37) {20};
\end{scope}
\begin{scope}
\path[clip] (  0.00,  0.00) rectangle (578.16,180.67);
\definecolor{drawColor}{RGB}{0,0,0}

\node[text=drawColor,anchor=base,inner sep=0pt, outer sep=0pt, scale=  1.40] at (305.97, 27.75) {Epoch};
\end{scope}
\begin{scope}
\path[clip] (  0.00,  0.00) rectangle (578.16,180.67);
\definecolor{drawColor}{RGB}{0,0,0}

\node[text=drawColor,rotate= 90.00,anchor=base,inner sep=0pt, outer sep=0pt, scale=  1.40] at (213.69,108.25) {Loss};
\end{scope}
\begin{scope}
\path[clip] (  0.00,  0.00) rectangle (578.16,180.67);
\definecolor{drawColor}{RGB}{0,0,0}

\node[text=drawColor,anchor=base,inner sep=0pt, outer sep=0pt, scale=  1.40] at (305.97,162.69) {\bfseries Loss};
\end{scope}
\begin{scope}
\path[clip] (429.12, 62.13) rectangle (571.16,154.36);
\definecolor{drawColor}{RGB}{255,255,255}

\path[draw=drawColor,line width= 0.7pt,line join=round] (429.12, 67.16) --
	(571.16, 67.16);

\path[draw=drawColor,line width= 0.7pt,line join=round] (429.12, 68.00) --
	(571.16, 68.00);

\path[draw=drawColor,line width= 0.7pt,line join=round] (429.12, 68.84) --
	(571.16, 68.84);

\path[draw=drawColor,line width= 0.7pt,line join=round] (429.12, 69.68) --
	(571.16, 69.68);

\path[draw=drawColor,line width= 0.7pt,line join=round] (429.12, 70.51) --
	(571.16, 70.51);

\path[draw=drawColor,line width= 0.7pt,line join=round] (429.12, 71.35) --
	(571.16, 71.35);

\path[draw=drawColor,line width= 0.7pt,line join=round] (429.12, 72.19) --
	(571.16, 72.19);

\path[draw=drawColor,line width= 0.7pt,line join=round] (429.12, 73.03) --
	(571.16, 73.03);

\path[draw=drawColor,line width= 0.7pt,line join=round] (429.12, 73.87) --
	(571.16, 73.87);

\path[draw=drawColor,line width= 0.7pt,line join=round] (429.12,142.62) --
	(571.16,142.62);

\path[draw=drawColor,line width= 0.7pt,line join=round] (429.12,143.46) --
	(571.16,143.46);

\path[draw=drawColor,line width= 0.7pt,line join=round] (429.12,144.30) --
	(571.16,144.30);

\path[draw=drawColor,line width= 0.7pt,line join=round] (429.12,145.14) --
	(571.16,145.14);

\path[draw=drawColor,line width= 0.7pt,line join=round] (429.12,145.98) --
	(571.16,145.98);

\path[draw=drawColor,line width= 0.7pt,line join=round] (429.12,146.82) --
	(571.16,146.82);

\path[draw=drawColor,line width= 0.7pt,line join=round] (429.12,147.66) --
	(571.16,147.66);

\path[draw=drawColor,line width= 0.7pt,line join=round] (429.12,148.49) --
	(571.16,148.49);

\path[draw=drawColor,line width= 0.7pt,line join=round] (429.12,149.33) --
	(571.16,149.33);

\path[draw=drawColor,line width= 0.7pt,line join=round] (436.86, 62.13) --
	(436.86,154.36);

\path[draw=drawColor,line width= 0.7pt,line join=round] (438.16, 62.13) --
	(438.16,154.36);

\path[draw=drawColor,line width= 0.7pt,line join=round] (439.45, 62.13) --
	(439.45,154.36);

\path[draw=drawColor,line width= 0.7pt,line join=round] (440.74, 62.13) --
	(440.74,154.36);

\path[draw=drawColor,line width= 0.7pt,line join=round] (442.03, 62.13) --
	(442.03,154.36);

\path[draw=drawColor,line width= 0.7pt,line join=round] (443.32, 62.13) --
	(443.32,154.36);

\path[draw=drawColor,line width= 0.7pt,line join=round] (444.61, 62.13) --
	(444.61,154.36);

\path[draw=drawColor,line width= 0.7pt,line join=round] (445.90, 62.13) --
	(445.90,154.36);

\path[draw=drawColor,line width= 0.7pt,line join=round] (447.19, 62.13) --
	(447.19,154.36);

\path[draw=drawColor,line width= 0.7pt,line join=round] (553.08, 62.13) --
	(553.08,154.36);

\path[draw=drawColor,line width= 0.7pt,line join=round] (554.37, 62.13) --
	(554.37,154.36);

\path[draw=drawColor,line width= 0.7pt,line join=round] (555.66, 62.13) --
	(555.66,154.36);

\path[draw=drawColor,line width= 0.7pt,line join=round] (556.96, 62.13) --
	(556.96,154.36);

\path[draw=drawColor,line width= 0.7pt,line join=round] (558.25, 62.13) --
	(558.25,154.36);

\path[draw=drawColor,line width= 0.7pt,line join=round] (559.54, 62.13) --
	(559.54,154.36);

\path[draw=drawColor,line width= 0.7pt,line join=round] (560.83, 62.13) --
	(560.83,154.36);

\path[draw=drawColor,line width= 0.7pt,line join=round] (562.12, 62.13) --
	(562.12,154.36);

\path[draw=drawColor,line width= 0.7pt,line join=round] (563.41, 62.13) --
	(563.41,154.36);
\definecolor{drawColor}{RGB}{211,211,211}

\path[draw=drawColor,line width= 0.7pt,line join=round] (429.12, 66.32) --
	(571.16, 66.32);

\path[draw=drawColor,line width= 0.7pt,line join=round] (429.12, 74.71) --
	(571.16, 74.71);

\path[draw=drawColor,line width= 0.7pt,line join=round] (429.12, 87.28) --
	(571.16, 87.28);

\path[draw=drawColor,line width= 0.7pt,line join=round] (429.12,108.25) --
	(571.16,108.25);

\path[draw=drawColor,line width= 0.7pt,line join=round] (429.12,129.21) --
	(571.16,129.21);

\path[draw=drawColor,line width= 0.7pt,line join=round] (429.12,141.79) --
	(571.16,141.79);

\path[draw=drawColor,line width= 0.7pt,line join=round] (429.12,150.17) --
	(571.16,150.17);

\path[draw=drawColor,line width= 0.7pt,line join=round] (435.57, 62.13) --
	(435.57,154.36);

\path[draw=drawColor,line width= 0.7pt,line join=round] (448.49, 62.13) --
	(448.49,154.36);

\path[draw=drawColor,line width= 0.7pt,line join=round] (467.86, 62.13) --
	(467.86,154.36);

\path[draw=drawColor,line width= 0.7pt,line join=round] (500.14, 62.13) --
	(500.14,154.36);

\path[draw=drawColor,line width= 0.7pt,line join=round] (532.42, 62.13) --
	(532.42,154.36);

\path[draw=drawColor,line width= 0.7pt,line join=round] (551.79, 62.13) --
	(551.79,154.36);

\path[draw=drawColor,line width= 0.7pt,line join=round] (564.70, 62.13) --
	(564.70,154.36);
\definecolor{drawColor}{RGB}{37,43,122}

\path[draw=drawColor,line width= 1.1pt,line join=round] (435.57, 66.32) --
	(436.35,143.46) --
	(436.99,146.48) --
	(437.12,147.15) --
	(437.51,147.74) --
	(438.03,148.07) --
	(438.03,147.99) --
	(438.03,147.91) --
	(438.03,147.74) --
	(438.16,148.33) --
	(438.28,148.33) --
	(438.41,148.49) --
	(438.41,148.41) --
	(438.80,148.58) --
	(438.93,148.58) --
	(439.06,148.58) --
	(439.32,148.91) --
	(439.32,148.58) --
	(440.22,149.16) --
	(441.25,149.25) --
	(445.90,149.67) --
	(564.70,150.17);
\definecolor{drawColor}{RGB}{211,211,211}

\path[draw=drawColor,line width= 0.6pt,line join=round] (429.12, 62.13) -- (571.16,154.36);
\end{scope}
\begin{scope}
\path[clip] (  0.00,  0.00) rectangle (578.16,180.67);
\definecolor{drawColor}{RGB}{0,0,0}

\path[draw=drawColor,line width= 0.6pt,line join=round,line cap=rect] (429.12, 62.13) --
	(429.12,154.36);
\end{scope}
\begin{scope}
\path[clip] (  0.00,  0.00) rectangle (578.16,180.67);
\definecolor{drawColor}{RGB}{0,0,0}

\node[text=drawColor,anchor=base east,inner sep=0pt, outer sep=0pt, scale=  0.50] at (422.62, 64.60) {0.00};

\node[text=drawColor,anchor=base east,inner sep=0pt, outer sep=0pt, scale=  0.50] at (422.62, 72.99) {0.10};

\node[text=drawColor,anchor=base east,inner sep=0pt, outer sep=0pt, scale=  0.50] at (422.62, 85.56) {0.25};

\node[text=drawColor,anchor=base east,inner sep=0pt, outer sep=0pt, scale=  0.50] at (422.62,106.52) {0.50};

\node[text=drawColor,anchor=base east,inner sep=0pt, outer sep=0pt, scale=  0.50] at (422.62,127.49) {0.75};

\node[text=drawColor,anchor=base east,inner sep=0pt, outer sep=0pt, scale=  0.50] at (422.62,140.06) {0.90};

\node[text=drawColor,anchor=base east,inner sep=0pt, outer sep=0pt, scale=  0.50] at (422.62,148.45) {1.00};
\end{scope}
\begin{scope}
\path[clip] (  0.00,  0.00) rectangle (578.16,180.67);
\definecolor{drawColor}{RGB}{0,0,0}

\path[draw=drawColor,line width= 0.6pt,line join=round] (425.62, 66.32) --
	(429.12, 66.32);

\path[draw=drawColor,line width= 0.6pt,line join=round] (425.62, 74.71) --
	(429.12, 74.71);

\path[draw=drawColor,line width= 0.6pt,line join=round] (425.62, 87.28) --
	(429.12, 87.28);

\path[draw=drawColor,line width= 0.6pt,line join=round] (425.62,108.25) --
	(429.12,108.25);

\path[draw=drawColor,line width= 0.6pt,line join=round] (425.62,129.21) --
	(429.12,129.21);

\path[draw=drawColor,line width= 0.6pt,line join=round] (425.62,141.79) --
	(429.12,141.79);

\path[draw=drawColor,line width= 0.6pt,line join=round] (425.62,150.17) --
	(429.12,150.17);
\end{scope}
\begin{scope}
\path[clip] (  0.00,  0.00) rectangle (578.16,180.67);
\definecolor{drawColor}{RGB}{0,0,0}

\path[draw=drawColor,line width= 0.6pt,line join=round,line cap=rect] (429.12, 62.13) --
	(571.16, 62.13);
\end{scope}
\begin{scope}
\path[clip] (  0.00,  0.00) rectangle (578.16,180.67);
\definecolor{drawColor}{RGB}{0,0,0}

\path[draw=drawColor,line width= 0.6pt,line join=round] (435.57, 58.63) --
	(435.57, 62.13);

\path[draw=drawColor,line width= 0.6pt,line join=round] (448.49, 58.63) --
	(448.49, 62.13);

\path[draw=drawColor,line width= 0.6pt,line join=round] (467.86, 58.63) --
	(467.86, 62.13);

\path[draw=drawColor,line width= 0.6pt,line join=round] (500.14, 58.63) --
	(500.14, 62.13);

\path[draw=drawColor,line width= 0.6pt,line join=round] (532.42, 58.63) --
	(532.42, 62.13);

\path[draw=drawColor,line width= 0.6pt,line join=round] (551.79, 58.63) --
	(551.79, 62.13);

\path[draw=drawColor,line width= 0.6pt,line join=round] (564.70, 58.63) --
	(564.70, 62.13);
\end{scope}
\begin{scope}
\path[clip] (  0.00,  0.00) rectangle (578.16,180.67);
\definecolor{drawColor}{RGB}{0,0,0}

\node[text=drawColor,anchor=base,inner sep=0pt, outer sep=0pt, scale=  0.50] at (435.57, 52.19) {0.00};

\node[text=drawColor,anchor=base,inner sep=0pt, outer sep=0pt, scale=  0.50] at (448.49, 52.19) {0.10};

\node[text=drawColor,anchor=base,inner sep=0pt, outer sep=0pt, scale=  0.50] at (467.86, 52.19) {0.25};

\node[text=drawColor,anchor=base,inner sep=0pt, outer sep=0pt, scale=  0.50] at (500.14, 52.19) {0.50};

\node[text=drawColor,anchor=base,inner sep=0pt, outer sep=0pt, scale=  0.50] at (532.42, 52.19) {0.75};

\node[text=drawColor,anchor=base,inner sep=0pt, outer sep=0pt, scale=  0.50] at (551.79, 52.19) {0.90};

\node[text=drawColor,anchor=base,inner sep=0pt, outer sep=0pt, scale=  0.50] at (564.70, 52.19) {1.00};
\end{scope}
\begin{scope}
\path[clip] (  0.00,  0.00) rectangle (578.16,180.67);
\definecolor{drawColor}{RGB}{0,0,0}

\node[text=drawColor,anchor=base,inner sep=0pt, outer sep=0pt, scale=  1.40] at (500.14, 27.75) {False positive fraction};
\end{scope}
\begin{scope}
\path[clip] (  0.00,  0.00) rectangle (578.16,180.67);
\definecolor{drawColor}{RGB}{0,0,0}

\node[text=drawColor,rotate= 90.00,anchor=base,inner sep=0pt, outer sep=0pt, scale=  1.40] at (406.41,108.25) {True positive fraction};
\end{scope}
\begin{scope}
\path[clip] (  0.00,  0.00) rectangle (578.16,180.67);
\definecolor{drawColor}{RGB}{0,0,0}

\node[text=drawColor,anchor=base,inner sep=0pt, outer sep=0pt, scale=  1.40] at (500.14,162.69) {\bfseries ROC Curve};
\end{scope}
\begin{scope}
\path[clip] (  0.00,  0.00) rectangle (578.16,180.67);
\definecolor{drawColor}{RGB}{0,0,0}

\node[text=drawColor,anchor=base west,inner sep=0pt, outer sep=0pt, scale=  1.40] at (193.71,  3.39) {Phase};
\end{scope}
\begin{scope}
\path[clip] (  0.00,  0.00) rectangle (578.16,180.67);
\definecolor{drawColor}{RGB}{134,182,178}

\path[draw=drawColor,line width= 1.7pt,line join=round] (242.00,  8.21) -- (253.56,  8.21);
\end{scope}
\begin{scope}
\path[clip] (  0.00,  0.00) rectangle (578.16,180.67);
\definecolor{drawColor}{RGB}{149,91,153}

\path[draw=drawColor,line width= 1.7pt,line join=round] (312.39,  8.21) -- (323.96,  8.21);
\end{scope}
\begin{scope}
\path[clip] (  0.00,  0.00) rectangle (578.16,180.67);
\definecolor{drawColor}{RGB}{0,0,0}

\node[text=drawColor,anchor=base west,inner sep=0pt, outer sep=0pt, scale=  1.20] at (262.01,  4.08) {Training};
\end{scope}
\begin{scope}
\path[clip] (  0.00,  0.00) rectangle (578.16,180.67);
\definecolor{drawColor}{RGB}{0,0,0}

\node[text=drawColor,anchor=base west,inner sep=0pt, outer sep=0pt, scale=  1.20] at (332.40,  4.08) {Validation};
\end{scope}
\end{tikzpicture}

%% file: writeup.bbl
\begin{thebibliography}{}

\bibitem[\protect\citeauthoryear{Boureau, Ponce \& LeCun}{Boureau
  et~al.}{2010}]{18}
Boureau, Y.-L., Ponce, J., \& LeCun, Y. (2010).
\newblock A theoretical analysis of feature pooling in visual recognition.
\newblock In {\em Proceedings of the 27th international conference on machine
  learning (ICML-10)}, (pp.\ 111--118).

\bibitem[\protect\citeauthoryear{Collobert, Kavukcuoglu \& Farabet}{Collobert
  et~al.}{2011}]{31}
Collobert, R., Kavukcuoglu, K., \& Farabet, C. (2011).
\newblock Torch7: A matlab-like environment for machine learning.
\newblock In {\em BigLearn, NIPS workshop}, number EPFL-CONF-192376.

\bibitem[\protect\citeauthoryear{Culotta, Ravi \& Cutler}{Culotta
  et~al.}{2016}]{42}
Culotta, A., Ravi, N.~K., \& Cutler, J. (2016).
\newblock Predicting twitter user demographics using distant supervision from
  website traffic data.
\newblock {\em Journal of Artificial Intelligence Research}, {\em 55},
  389--408.

\bibitem[\protect\citeauthoryear{dos Santos \& Gatti}{dos Santos \&
  Gatti}{2014}]{5}
dos Santos, C. \& Gatti, M. (2014).
\newblock Deep convolutional neural networks for sentiment analysis of short
  texts.
\newblock In {\em Proceedings of COLING 2014, the 25th International Conference
  on Computational Linguistics: Technical Papers}, (pp.\ 69--78).

\bibitem[\protect\citeauthoryear{Frome, Corrado, Shlens, Bengio, Dean, Mikolov
  \& others}{Frome et~al.}{2013}]{23}
Frome, A., Corrado, G.~S., Shlens, J., Bengio, S., Dean, J., Mikolov, T.,
  et~al. (2013).
\newblock Devise: A deep visual-semantic embedding model.
\newblock In {\em Advances in neural information processing systems}, (pp.\
  2121--2129).

\bibitem[\protect\citeauthoryear{Gao, He, Yih \& Deng}{Gao et~al.}{2013}]{24}
Gao, J., He, X., Yih, W.-t., \& Deng, L. (2013).
\newblock Learning semantic representations for the phrase translation model.
\newblock {\em arXiv preprint arXiv:1312.0482}.

\bibitem[\protect\citeauthoryear{Gawrilow, Langevin, Jonker, Coppersmith,
  Hilland, Morgan \& Azunre}{Gawrilow et~al.}{2018}]{41}
Gawrilow, J., Langevin, S., Jonker, D.and~Bethune, C., Coppersmith, G.,
  Hilland, C., Morgan, J., \& Azunre, P. (2018).
\newblock Distil: A mixed-initiative model discovery system for subject matter
  experts.
\newblock In {\em Proceedings of the International Workshop on Automatic
  Machine Learning at ICML}, volume~2.

\bibitem[\protect\citeauthoryear{Girshick, Donahue, Darrell \& Malik}{Girshick
  et~al.}{2014}]{28}
Girshick, R., Donahue, J., Darrell, T., \& Malik, J. (2014).
\newblock Rich feature hierarchies for accurate object detection and semantic
  segmentation.
\newblock In {\em Proceedings of the IEEE conference on computer vision and
  pattern recognition}, (pp.\ 580--587).

\bibitem[\protect\citeauthoryear{Graves}{Graves}{2012}]{19}
Graves, A. (2012).
\newblock {\em Supervised sequence labelling with recurrent neural networks},
  volume 385.
\newblock Studies in Computational Intelligence. Springer, Berlin, Heidelberg.

\bibitem[\protect\citeauthoryear{Hochreiter \& Schmidhuber}{Hochreiter \&
  Schmidhuber}{1997}]{20}
Hochreiter, S. \& Schmidhuber, J. (1997).
\newblock Long short-term memory.
\newblock {\em Neural computation}, {\em 9\/}(8), 1735--1780.

\bibitem[\protect\citeauthoryear{Joachims}{Joachims}{1998}]{3}
Joachims, T. (1998).
\newblock Text categorization with support vector machines: Learning with many
  relevant features.
\newblock {\em Proceedings of the 10th European Conferenece on Machine
  Learning, 1998}, 137--142.

\bibitem[\protect\citeauthoryear{Johnson \& Zhang}{Johnson \& Zhang}{2015}]{14}
Johnson, R. \& Zhang, T. (2015).
\newblock Effective use of word order for text categorization with
  convolutional neural networks.
\newblock In {\em Proceedings of the 2015 Conference of the North American
  Chapter of the Association for Computational Linguistics: Human Language
  Technologies}, (pp.\ 103--112).

\bibitem[\protect\citeauthoryear{Kalchbrenner, Grefenstette \&
  Blunsom}{Kalchbrenner et~al.}{2014}]{16}
Kalchbrenner, N., Grefenstette, E., \& Blunsom, P. (2014).
\newblock A convolutional neural network for modelling sentences.
\newblock {\em Proceedings of the 52nd Annual Meeting of the Association for
  Computational Linguistics}.

\bibitem[\protect\citeauthoryear{Kanaris, Kanaris, Houvardas \&
  Stamatatos}{Kanaris et~al.}{2007}]{10}
Kanaris, I., Kanaris, K., Houvardas, I., \& Stamatatos, E. (2007).
\newblock Words versus character n-grams for anti-spam filtering.
\newblock {\em International Journal on Artificial Intelligence Tools}, {\em
  16\/}(06), 1047--1067.

\bibitem[\protect\citeauthoryear{Kim}{Kim}{2014}]{4}
Kim, Y. (2014).
\newblock Convolutional neural networks for sentence classification.
\newblock In {\em Proceedings of the 2014 Conference on Empirical Methods in
  Natural Language Processing (EMNLP)}, (pp.\ 1746--1751).

\bibitem[\protect\citeauthoryear{Kim, Stratos, Sarikaya \& Jeong}{Kim
  et~al.}{2015}]{40}
Kim, Y.-B., Stratos, K., Sarikaya, R., \& Jeong, M. (2015).
\newblock New transfer learning techniques for disparate label sets.
\newblock In {\em Proceedings of the 53rd Annual Meeting of the Association for
  Computational Linguistics and the 7th International Joint Conference on
  Natural Language Processing (Volume 1: Long Papers)}, volume~1, (pp.\
  473--482).

\bibitem[\protect\citeauthoryear{Le \& Mikolov}{Le \& Mikolov}{2014}]{25}
Le, Q. \& Mikolov, T. (2014).
\newblock Distributed representations of sentences and documents.
\newblock In {\em International Conference on Machine Learning}, (pp.\
  1188--1196).

\bibitem[\protect\citeauthoryear{LeCun, Bottou, Bengio \& Haffner}{LeCun
  et~al.}{1998}]{30}
LeCun, Y., Bottou, L., Bengio, Y., \& Haffner, P. (1998).
\newblock Gradient-based learning applied to document recognition.
\newblock {\em Proceedings of the IEEE}, {\em 86\/}(11), 2278--2324.

\bibitem[\protect\citeauthoryear{Lilleberg, Zhu \& Zhang}{Lilleberg
  et~al.}{2015}]{11}
Lilleberg, J., Zhu, Y., \& Zhang, Y. (2015).
\newblock Support vector machines and word2vec for text classification with
  semantic features.
\newblock In {\em Cognitive Informatics \& Cognitive Computing (ICCI* CC), 2015
  IEEE 14th International Conference on}, (pp.\ 136--140). IEEE.

\bibitem[\protect\citeauthoryear{Liu, Pasupat, Wang, Cyphers \& Glass}{Liu
  et~al.}{2013}]{38}
Liu, J., Pasupat, P., Wang, Y., Cyphers, S., \& Glass, J. (2013).
\newblock Query understanding enhanced by hierarchical parsing structures.
\newblock In {\em Automatic Speech Recognition and Understanding (ASRU), 2013
  IEEE Workshop on}, (pp.\ 72--77). Citeseer.

\bibitem[\protect\citeauthoryear{Mikolov, Le \& Sutskever}{Mikolov
  et~al.}{2013}]{26}
Mikolov, T., Le, Q.~V., \& Sutskever, I. (2013).
\newblock
\newblock Exploiting similarities among languages for machine translation.

\bibitem[\protect\citeauthoryear{Mikolov, Sutskever, Chen, Corrado \&
  Dean}{Mikolov et~al.}{2013}]{12}
Mikolov, T., Sutskever, I., Chen, K., Corrado, G.~S., \& Dean, J. (2013).
\newblock Distributed representations of words and phrases and their
  compositionality.
\newblock In {\em Advances in neural information processing systems}, (pp.\
  3111--3119).

\bibitem[\protect\citeauthoryear{Mohri, Rostamizadeh \& Talwalkar}{Mohri
  et~al.}{2018}]{15}
Mohri, M., Rostamizadeh, A., \& Talwalkar, A. (2018).
\newblock {\em Foundations of machine learning}.
\newblock MIT press.

\bibitem[\protect\citeauthoryear{Mou, Meng, Yan, Li, Xu, Zhang \& Jin}{Mou
  et~al.}{2016}]{39}
Mou, L., Meng, Z., Yan, R., Li, G., Xu, Y., Zhang, L., \& Jin, Z. (2016).
\newblock How transferable are neural networks in nlp applications?
\newblock In {\em Proceedings of the 2016 Conference on Empirical Methods in
  Natural Language Processing}, (pp.\ 479--489).

\bibitem[\protect\citeauthoryear{Pennington, Socher \& Manning}{Pennington
  et~al.}{2014}]{27}
Pennington, J., Socher, R., \& Manning, C. (2014).
\newblock Glove: Global vectors for word representation.
\newblock In {\em Proceedings of the 2014 conference on empirical methods in
  natural language processing (EMNLP)}, (pp.\ 1532--1543).

\bibitem[\protect\citeauthoryear{Ritter, Clark, Etzioni \& others}{Ritter
  et~al.}{2011}]{37}
Ritter, A., Clark, S., Etzioni, O., et~al. (2011).
\newblock Named entity recognition in tweets: an experimental study.
\newblock In {\em Proceedings of the conference on empirical methods in natural
  language processing}, (pp.\ 1524--1534). Association for Computational
  Linguistics.

\bibitem[\protect\citeauthoryear{Rodriguez, Caldwell \& Liu}{Rodriguez
  et~al.}{2018}]{36}
Rodriguez, J.~D., Caldwell, A., \& Liu, A. (2018).
\newblock Transfer learning for entity recognition of novel classes.
\newblock In {\em Proceedings of the 27th International Conference on
  Computational Linguistics}, (pp.\ 1974--1985).

\bibitem[\protect\citeauthoryear{Santos \& Zadrozny}{Santos \&
  Zadrozny}{2014}]{8}
Santos, C.~D. \& Zadrozny, B. (2014).
\newblock Learning character-level representations for part-of-speech tagging.
\newblock In {\em Proceedings of the 31st International Conference on Machine
  Learning (ICML-14)}, (pp.\ 1818--1826).

\bibitem[\protect\citeauthoryear{Shen, He, Gao, Deng \& Mesnil}{Shen
  et~al.}{2014}]{7}
Shen, Y., He, X., Gao, J., Deng, L., \& Mesnil, G. (2014).
\newblock A latent semantic model with convolutional-pooling structure for
  information retrieval.
\newblock In {\em Proceedings of the 23rd ACM International Conference on
  Conference on Information and Knowledge Management}, (pp.\ 101--110). ACM.

\bibitem[\protect\citeauthoryear{Soderland}{Soderland}{2001}]{1}
Soderland, S. (2001).
\newblock Building a machine learning based text understanding system.
\newblock In {\em In Proc. IJCAI-2001 Workshop on Adaptive Text Extraction and
  Mining}, (pp.\ 64--70).

\bibitem[\protect\citeauthoryear{Srivastava, Hinton, Krizhevsky, Sutskever \&
  Salakhutdinov}{Srivastava et~al.}{2014}]{17}
Srivastava, N., Hinton, G., Krizhevsky, A., Sutskever, I., \& Salakhutdinov, R.
  (2014).
\newblock Dropout: a simple way to prevent neural networks from overfitting.
\newblock {\em The Journal of Machine Learning Research}, {\em 15\/}(1),
  1929--1958.

\bibitem[\protect\citeauthoryear{Wang \& Domeniconi}{Wang \&
  Domeniconi}{2008}]{13}
Wang, P. \& Domeniconi, C. (2008).
\newblock Building semantic kernels for text classification using wikipedia.
\newblock In {\em Proceedings of the 14th ACM SIGKDD international conference
  on Knowledge discovery and data mining}, (pp.\ 713--721). ACM.

\bibitem[\protect\citeauthoryear{Yang, Li, Ding \& Li}{Yang et~al.}{2013}]{9}
Yang, L., Li, C., Ding, Q., \& Li, L. (2013).
\newblock Combining lexical and semantic features for short text
  classification.
\newblock {\em Procedia Computer Science}, {\em 22}, 78--86.

\bibitem[\protect\citeauthoryear{Zhang, Zhao \& LeCun}{Zhang et~al.}{2015}]{2}
Zhang, X., Zhao, J., \& LeCun, Y. (2015).
\newblock Character-level convolutional networks for text classification.
\newblock In {\em Proceedings of the 28th International Conference on Neural
  Information Processing Systems - Volume 1}, NIPS'15, (pp.\ 649--657).,
  Cambridge, MA, USA. MIT Press.

\end{thebibliography}
